\documentclass{article}
\usepackage{PRIMEarxiv}
\usepackage{algorithm}
\usepackage{algpseudocode}

\usepackage[english]{babel}
\addto\captionsenglish{
	
}

\usepackage{hyperref}       
\usepackage{url}            
\usepackage{amsfonts}  
\usepackage{enumerate}
\usepackage{verbatim}
\usepackage{amsmath,amsthm}
\usepackage{multirow}
\usepackage{epsfig}
\usepackage{booktabs}
\usepackage{graphicx} 
\usepackage{subcaption}
\usepackage{makecell}
\usepackage{colortbl}  
\usepackage{xcolor}
\usepackage{array}   
\usepackage{caption}
\usepackage{cleveref}
\crefname{figure}{Fig.}{Figs.}

\usepackage{amsmath}   
\usepackage{amssymb}   

\usepackage{titlesec}
\usepackage{bbding}

\usepackage{makecell}
\usepackage{rotating}

\titleformat{\paragraph}
{\normalfont\normalsize\bfseries}{\theparagraph}{0.5em}{}
\titlespacing*{\paragraph}
{0pt}{3.25ex plus 1ex minus .2ex}{1.5ex plus .2ex}

\usepackage[utf8]{inputenc} 
\usepackage[T1]{fontenc}    
\usepackage{hyperref}       
\usepackage{url}            
\usepackage{booktabs}       
\usepackage{amsfonts}       
\usepackage{nicefrac}       
\usepackage{microtype}      
\usepackage{lipsum}
\usepackage{fancyhdr}       
\usepackage{graphicx}       
\graphicspath{{media/}}     

\title{Semantic-Deviation–Anchored Multi-Branch Fusion for Unsupervised Anomaly Detection and Localization in Unstructured Conveyor-Belt Coal Scenes
}

\author{
  Wenping Jin, Yuyang Tang, Li Zhu \\
  School of Software Engineering / School of Mechanical Engineering \\
  Xi'an Jiaotong University \\
  Xi'an, China\\
  \texttt{jinwenping@stu.xjtu.edu.cn, tangyuyang@xjtu.edu.cn, zhuli@xjtu.edu.cn} \\
}

\begin{document}
\maketitle

\begin{abstract}
	Reliable foreign-object anomaly detection and pixel-level localization in conveyor-belt coal scenes are essential for safe and intelligent mining operations. This task is particularly challenging due to the highly unstructured environment: coal and gangue are randomly piled, backgrounds are complex and variable, and foreign objects often exhibit low contrast, deformation, occlusion, resulting in coupling with their surroundings. These characteristics weaken the stability and regularity assumptions that many anomaly detection methods rely on in structured industrial settings, leading to notable performance degradation. To support evaluation and comparison in this setting, we construct \textbf{CoalAD}, a benchmark for unsupervised foreign-object anomaly detection with pixel-level localization in coal-stream scenes. We further propose a complementary-cue collaborative perception framework that extracts and fuses complementary anomaly evidence from three perspectives: object-level semantic composition modeling, semantic-attribution-based global deviation analysis, and fine-grained texture matching. The fused outputs provide robust image-level anomaly scoring and accurate pixel-level localization. Experiments on CoalAD demonstrate that our method outperforms widely used baselines across the evaluated image-level and pixel-level metrics, and ablation studies validate the contribution of each component. The code is available at \url{https://github.com/xjpp2016/USAD}.
\end{abstract}


\section{Introduction}

In coal mining operations, coal and gangue are typically extracted together and continuously transported in a mixed form on conveyor belts. Under complex excavation and transportation conditions, various types of production debris and foreign objects---such as wood pieces, metal components, and plastic products---can easily be mixed into the coal stream. These intrusions not only reduce the efficiency of downstream screening and washing, but may also cause jamming, abrasion, or even structural damage to critical equipment, potentially leading to unplanned downtime and safety incidents~\cite{DsCGF}. Therefore, reliable intelligent detection and pixel-level localization of foreign objects in conveyor-belt coal scenes is crucial for production safety and the advancement of intelligent mining. Fundamentally, this task involves two coupled objectives: (1) determining whether a foreign object exists (\emph{detection}), and (2) precisely describing its location and shape (\emph{segmentation}). Together, these outputs provide essential perceptual evidence for downstream automated decision-making and execution, such as guiding a robotic arm to remove intrusions.

In real-world industrial environments, the occurrence of foreign objects is highly sporadic and unpredictable, and their shapes, sizes, and appearances vary substantially, making it difficult to pre-enumerate all categories. As a result, supervised learning methods that rely on large-scale annotated foreign-object data struggle to cover the full spectrum of real operating conditions. In contrast, unsupervised methods that model normality using only normal samples are inherently more suitable for handling unknown or rare intrusions. This setting aligns with a long-standing research direction in industrial vision: anomaly detection and segmentation (often referred to as anomaly localization).

\begin{figure}[t]
	\centering
	\includegraphics[width=0.95\linewidth]{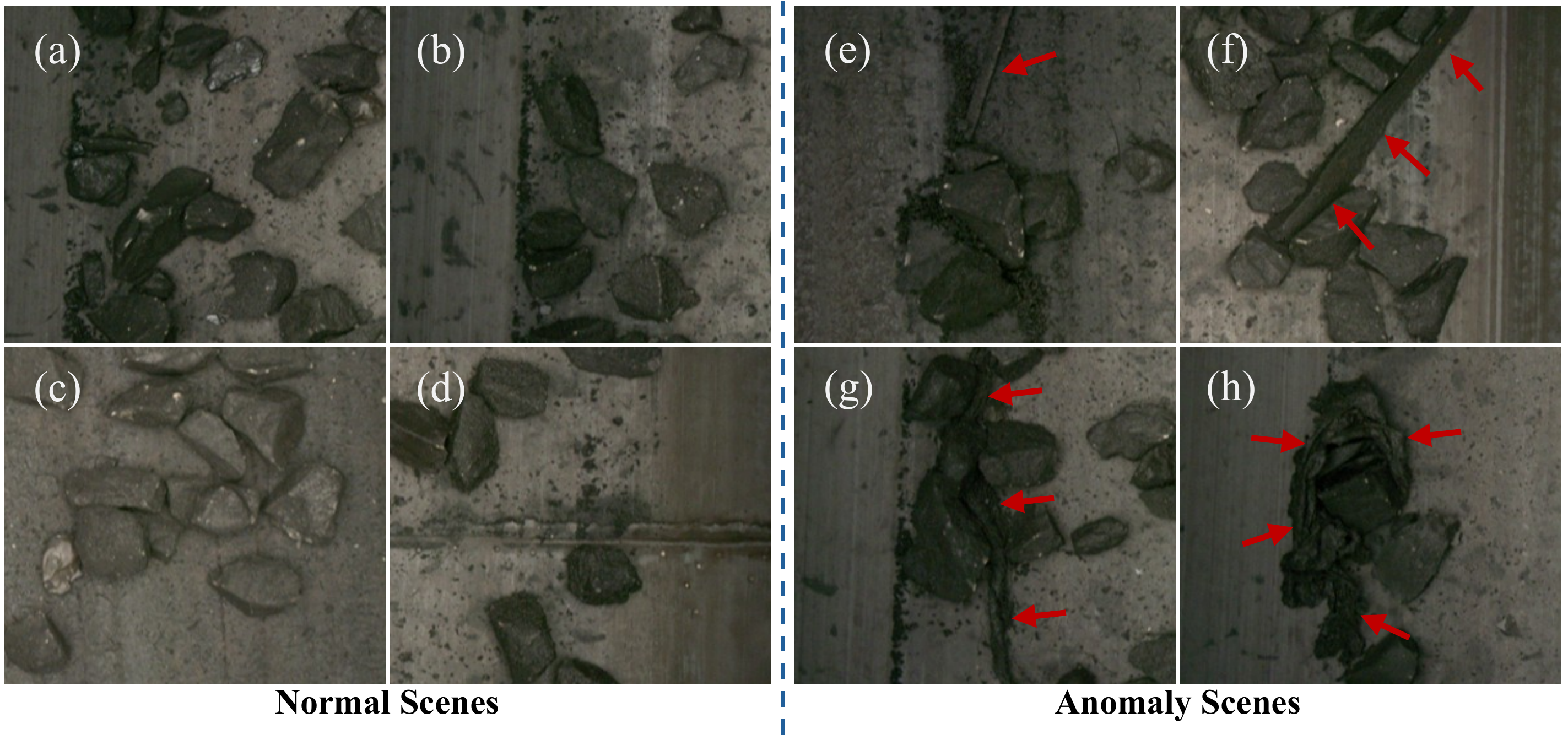}
	\caption{\textbf{Unstructured characteristics and low-contrast anomaly examples in conveyor-belt coal-stream scenes.} Normal samples are shown on the left (a--d) and anomalous samples on the right (e--h). Normal scenes exhibit randomly piled and intermixed coal and gangue with irregular sizes, shapes, and spatial distributions; meanwhile, belt wear patterns and coal dust introduce complex backgrounds. In anomalous scenes, foreign objects (e.g., wood, nets and/or ropes, and bags) are tightly coupled with the coal stream and thus hard to distinguish due to low contrast, occlusion, and discoloration, often yielding blurred boundaries.}
	\label{fig:Coal_datafirst}
\end{figure}

In recent years, a series of competitive, representative methods have been proposed for industrial anomaly detection and localization, e.g., PatchCore~\cite{PatchCore}, DRAEM~\cite{DRAEM}, RD4AD/RD++~\cite{RD4AD,RDplus}, EfficientAD~\cite{EfficientAD}, and SimpleNet~\cite{SimpleNet}, representing memory-based, reconstruction-based, reverse-distillation, student–teacher, and synthesis-driven paradigms. Although these methods achieve strong performance in structured industrial scenarios, their effectiveness often relies on implicit assumptions, such as stable object appearance, regular layouts, and limited background variation. In contrast, conveyor-belt coal scenes constitute a \textbf{highly unstructured environment}, where these assumptions are frequently violated. As shown in Fig.~\ref{fig:Coal_datafirst}, coal and gangue are randomly piled with continuously varying particle sizes; the belt background exhibits complex textures with wear and stains; and additional disturbances such as dust, illumination fluctuations, and occlusions make the \emph{normal pattern} itself highly diverse and time-varying. More importantly, unlike typical industrial defect scenarios where anomalies manifest as surface defects or structural damage with clear contrast against normal regions, \textbf{foreign objects} in coal streams (e.g., wooden rods, nets, ropes, bagged items) are often deformed by compression, partially occluded, or discolored, leading to \textbf{strong coupling with the coal background} in color, shape, and texture. Consequently, anomaly boundaries become blurred and the contrast is often low—and in a subset of cases, very low—making anomalies easily confused with normal coal blocks, gangue, and complex backgrounds. To advance research in this realistic and challenging setting, we construct a dataset, termed \textbf{CoalAD}\footnote{\url{https://www.modelscope.cn/datasets/lyfjwp/CoalAD}}, for unsupervised foreign-object anomaly detection and pixel-level localization in conveyor-belt coal scenes, and conduct a systematic evaluation of representative baselines on this benchmark. The results indicate that, due to the highly unstructured nature and strong disturbances of this scenario, the performance of existing methods degrades substantially (Table~\ref{tab:auc_comparison_with_venue}).

From a methodological perspective, many classical paradigms designed for structured industrial scenarios follow a \emph{bottom-up} accumulation of anomaly evidence. They assume that local normal patterns are relatively stable, while anomalies typically appear as local structural corruption or deviation. Accordingly, memory-bank-based methods detect anomalies by comparing local test features to normal features stored in the memory bank; reconstruction-based methods treat the ability to reconstruct local regions under the normal model as evidence of normality; and distillation- or teacher--student-based methods detect deviations by enforcing consistency of multi-layer local features between teacher and student networks. However, in conveyor-belt coal scenes, \textbf{local regions themselves are highly unstructured and stochastic}: drastic local appearance changes do not necessarily indicate anomalies; conversely, some foreign objects may resemble normal coal in local texture or color, causing \textbf{purely local similarity-based criteria to fail}. Therefore, this task requires introducing more interpretable anomaly cues at higher semantic levels and robustly propagating global/semantic anomaly indications to anomaly localization.

Based on the above analysis, we argue that unsupervised anomaly detection and localization for conveyor-belt coal scenes should be built upon a \textbf{multi-level cooperative perception and reasoning framework}. For detection, global semantic deviation provides a more robust anomaly criterion, while local cues serve as complementary evidence. For localization, anomaly regions should be jointly constrained from \textbf{three complementary perspectives: (1) object-level modeling to detect and localize anomalous objects; (2) associating global semantic anomaly signals with spatial positions to support localization decisions; and (3) leveraging fine-grained texture differences as additional discriminative cues}. Through their synergy, localization robustness can be improved under low contrast and heavy occlusion.

Specifically, for anomaly localization, we propose a three-branch complementary framework that extracts anomaly cues from object-level, semantic-level (via attribution analysis), and texture-level perspectives. The object-level and semantic-level branches adopt DINOv2~\cite{DINOv2}, a pre-trained Vision Transformer (ViT)~\cite{ViT} with strong semantic representations and good cross-domain generalization, as the backbone. In the object-level branch, patch tokens from normal samples are clustered to establish statistical distributions for background (conveyor belt) and foreground (coal and gangue). During inference, regions that do not belong to either the learned background or foreground distributions are identified as anomalous object regions. The semantic-level branch performs contribution analysis: an ablation-based strategy is used to estimate each patch's contribution to the global semantic anomaly, which is then exploited for localization. The texture-level branch uses a ResNet-based CNN backbone~\cite{ResNet} and performs anomaly discrimination in a PatchCore-like manner. Finally, we fuse multi-branch evidence to refine localization, using complementary cues to suppress spurious responses and produce the final anomaly maps.

Furthermore, for image-level anomaly detection, we integrate global semantic deviation with localization-aware cues. We first fit a Gaussian model to DINOv2 global features (\texttt{[CLS]} tokens) on normal samples and use the distance as the primary score. On top of this, we add two local components: (1) a spatial cue aggregated from the fused localization map, and (2) the texture-branch anomaly score. The three components are combined to form the final image-level anomaly score.

Our main contributions are summarized as follows:
\begin{itemize}
	\item \textbf{CoalAD benchmark.}
	We curate and release \textbf{CoalAD} for unsupervised foreign-object anomaly detection and pixel-level localization in unstructured conveyor-belt coal scenes (2,490 normal training images; 1,754 test images with 943 anomalous), with pixel-level ground-truth masks for evaluation.
	
	\item \textbf{Multi-cue localization with object composition, semantic attribution, and texture evidence.}
	We design a multi-cue localization framework with three complementary branches: an object-composition branch (foreground--background modeling via token clustering), a semantic attribution branch that decomposes global semantic deviation into patch-wise contributions via an efficient closed-form ablation, and a texture branch instantiated with PatchCore-style nearest-neighbor matching. The three cues are fused to produce the final anomaly localization maps.

	\item \textbf{Global semantics with localization-aware scoring.}
	We anchor image-level anomaly detection on global semantic deviation and enhance it with localization-aware evidence aggregated from predicted anomaly maps.
	
	\item \textbf{Experimental validation.}
	We conduct comprehensive experiments on CoalAD, including comparisons with representative baselines and ablation studies, to validate the effectiveness of the proposed components.
\end{itemize}

\section{Related Works}
\subsection{Industrial Visual Anomaly Detection and Localization Datasets}

Industrial visual anomaly detection typically follows the setting of ``training with normal samples only and detecting/localizing anomalies at test time'', which addresses practical challenges in real production lines where anomalous samples are scarce and their appearances are hard to anticipate. The most widely used benchmark dataset is \textbf{MVTec AD}~\cite{MVTecAD}. It covers diverse industrial products and provides pixel-level annotations; its anomalies mainly correspond to typical patterns such as structural defects and texture damage. Based on this benchmark, researchers have developed representative paradigms including memory-bank matching, reconstruction/generation, teacher--student distillation, and synthesis-driven discriminative learning, leading to substantial performance improvements. Meanwhile, datasets such as \textbf{VisA}~\cite{VisA} and \textbf{MPDD}~\cite{MPDD} further expand scenario coverage and defect types, increasing the diversity of samples and defect patterns. In recent years, the community has released larger-scale and more complex benchmarks, such as \textbf{Real-IAD} (multi-view, large-scale)~\cite{REAL-IAD,REAL-IAD-V} and \textbf{MANTA} (tiny objects, multi-view)~\cite{MANTA}. These datasets place more emphasis on simulating view changes, defect diversity, and distribution shifts in real industrial environments, thereby imposing higher requirements on model generalization.

Beyond the above paradigms that mainly target structural or texture anomalies, \textbf{MVTec LOCO AD}~\cite{LOCO-AD} further introduces a more challenging task of logical anomaly detection. Under this setting, models must judge whether a test image is semantically and logically consistent with normal samples (e.g., category, quantity, and placement), rather than only being consistent in low-level appearance or structure. Consequently, many methods that perform strongly on structural anomaly detection exhibit varying degrees of performance degradation under the LOCO setting, which has also spurred a line of methods dedicated to logical anomaly detection~\cite{EfficientAD,ComAD, SALAD}.

It is worth noting that the overall scenes of the above mainstream benchmarks remain largely highly structured: normal samples are typically highly consistent in terms of number, shape, and layout. This is fundamentally different from the unstructured scenario considered in this work. In unstructured environments, the lack of stable geometric layouts and appearance priors breaks many implicit assumptions behind structured settings, often leading to noticeable performance degradation.

\subsection{Metric-based Anomaly Detection and Localization in Feature Space}

Our method essentially belongs to the metric-based paradigm for anomaly detection in feature space. The core idea is to directly measure how far a test sample deviates from the learned normal patterns in the feature space, which is one of the dominant approaches in this field. Specifically, a visual encoder is used to extract feature representations from the whole image or local patches, which are then compared against a pre-defined normal reference to compute distances or deviations, resulting in image-level and pixel-level anomaly scores. Depending on how the normal reference is represented in feature space (e.g., instance sets, parametric statistics, or boundary constraints), metric-based methods can be grouped into three categories.

The first category is \textbf{instance-based matching with normal exemplars}. 
Instead of fitting an explicit parametric model, it uses features extracted from normal training data as exemplar references. At test time, anomaly scores are computed by nearest-neighbor matching (e.g., kNN~\cite{KNN}) between test features and these normal exemplars. Representative works include SPADE~\cite{SPADE}, PatchCore~\cite{PatchCore} and PNI~\cite{PNI}. This line is simple to implement and efficient to train, but it often focuses on local patterns and can be sensitive to exemplar coverage as well as retrieval/storage scalability.

The second category is \textbf{distribution-based parametric modeling}. 
It assumes that normal features follow a parameterized distribution (e.g., a multivariate Gaussian), and quantifies the deviation of test features from this distribution using statistics such as the Mahalanobis distance. GaussianAD~\cite{GaussianAD} and PaDiM~\cite{PaDiM} are typical examples. These methods are memory-efficient and interpretable, and can be sensitive to semantic anomalies (e.g., logical anomalies)~\cite{PUAD}. However, their performance depends heavily on the feature extraction capability of the encoder; when the representation is insufficient, the simple distributional assumption may lead to distorted or invalid measurements.

The third category is \textbf{one-class boundary or hypersphere constraints}. 
It aims to learn a compact decision boundary that tightly encloses the normal feature space, for example by extending support vector data description (SVDD)~\cite{SVDD} to patch-level features~\cite{PatchSVDD} and using the distance to the boundary or normal feature center as the anomaly criterion. Such methods impose an explicit compactness constraint on the normal feature space and use the distance to the learned center/hypersphere (or decision function) as the anomaly score; however, their objectives are often relatively simple and may overfit to training normal data, potentially weakening the representation’s discriminability. 

Our proposed method combines a \emph{parametric semantic reference} (distribution modeling) with a \emph{non-parametric local reference} (instance matching): the former captures high-level semantic deviations with a strong encoder (e.g., DINOv2~\cite{DINOv2}), while the latter remains sensitive to fine-grained texture/structural anomalies, leading to more comprehensive and robust detection.

\subsection{Reconstruction/Generation-based and Distillation-based Methods}

In addition to distance-based methods, our experiments also compare against representative works from two other mainstream families: reconstruction/generation methods and teacher--student distillation methods.

\textbf{Reconstruction/generation methods} aim to model the distribution of normal data by learning to reconstruct or generate normal samples, and use reconstruction error, feature residuals, or predictive uncertainty as anomaly cues at test time~\cite{RES-AD2017,RES-AD2018,RES-AD2019}. Typical implementations rely on generative models such as autoencoders (AE)~\cite{AE}, variational autoencoders (VAE)~\cite{VAE}, and generative adversarial networks (GAN)~\cite{GAN}. The underlying motivation is that a model trained only on normal samples has limited ability to reconstruct or generate anomalous patterns, which leads to noticeable reconstruction deviations at the pixel level and thus enables anomaly localization and detection.

However, in practice, reconstruction models often suffer from over-generalization: even for anomalous regions, the model may still produce seemingly plausible reconstructions~\cite{NotRecAD}. To alleviate this issue, various improvements have been proposed. For example, DRAEM~\cite{DRAEM} enhances anomaly sensitivity by introducing synthetic defects; ISSTAD~\cite{ISSTAD} upgrades the backbone to a ViT to improve representation capacity; other studies explore stronger generators such as diffusion models to better fit the normal data distribution~\cite{DFAD1,DFAD2,DFAD3}. Nevertheless, the effectiveness of reconstruction-based paradigms still depends largely on the structural stability of normal samples and the learnability of their distribution. In highly unstructured scenes with drastic background or appearance variations, it may be difficult to learn stable reconstruction regularities, leading to degraded performance.

\textbf{Distillation-based methods} focus on learning stable normal feature representations from normal data. They typically adopt a teacher--student framework, where the teacher network provides target features or intermediate representations and the student network is trained to mimic the teacher outputs~\cite{KD2015}. At test time, the feature discrepancy between the student and teacher is used as an anomaly measure. Compared with reconstruction methods, teacher--student approaches rely more on feature-level consistency than pixel-level generation quality, and can thus better exploit high-level semantic features to characterize normal patterns. For instance, EfficientAD~\cite{EfficientAD} performs well in logical anomaly detection scenes that require modeling semantic consistency.

Beyond the conventional teacher--student architecture with forward information transfer, industrial anomaly detection also includes a strong variant known as reverse distillation. It increases the learning difficulty of the student by adopting a reverse feature transfer mechanism, thereby strengthening its modeling of normal patterns and sensitivity to anomalies. Representative methods include RD4AD~\cite{RD4AD} and RD$^{++}$~\cite{RDplus}.

Despite their effectiveness, teacher--student architectures have clear limitations: the student tends to imitate the teacher outputs rather than truly understand semantic content, which weakens fine-grained semantic discrimination. In particular, when anomalies have low separability from the background, the student may almost perfectly replicate the teacher responses, resulting in degraded discriminative performance.

\section{Dataset}
\label{sec:dataset}

\subsection{Data Source and Curation}
Our experimental data are curated from the public DsCGF dataset released by Lv et al.~\cite{DsCGF}, which was collected in a real coal preparation plant. To construct an unsupervised anomaly detection benchmark tailored for online sorting and safety monitoring, we reformulate DsCGF as follows. We select the subset containing foreign objects, captured from an overhead conveyor viewpoint, and perform preprocessing steps including removal of duplicate scenes, splitting into normal and anomalous subsets, and resizing to a unified resolution. The resulting dataset is termed \textbf{CoalAD}.

\subsection{Task Setting and Characteristics}
In CoalAD, models are trained using only normal samples (containing only coal and gangue), and are required to detect and localize at the pixel level test samples containing unknown foreign objects (e.g., wooden debris, metal parts, nets, ropes, plastic bags).
This setting is motivated by the practical characteristics of conveyor-belt coal scenes: foreign objects occur sporadically, their categories are open-ended, and their appearance can change dramatically due to compression, occlusion, or coal-dust contamination.
As a result, collecting exhaustive anomalous categories and obtaining dense pixel-level annotations are costly and often infeasible, making fully supervised anomaly detection difficult to deploy in practice.

Meanwhile, CoalAD remains highly challenging even under the normal-only training protocol.
The coal stream and conveyor background are strongly non-structural and highly disturbed, and anomalous regions are often tightly coupled with the background (e.g., partially buried, heavily contaminated, or visually similar to coal textures), leading to weak boundaries and low contrast (see Fig.~\ref{fig:Coal_datafirst}).
This presents a sharp contrast to structured industrial benchmarks such as MVTec AD, which typically assume regular layouts and stable appearances, and where anomalies are relatively well-separated from backgrounds.

\subsection{Annotation and Ground-Truth Preparation}
To enable pixel-level localization evaluation, we generate binary anomaly masks for all anomalous samples by leveraging the instance segmentation annotations in DsCGF.
All foreign-object instances are merged and labeled as foreground (value 1), while the remaining regions are labeled as background (value 0).
We additionally conduct careful manual review to consolidate the masks (e.g., boundary clarification and consistency checking), aiming to provide reliable ground truth for evaluation.

\subsection{Data Splits and Statistics}
\begin{figure}[t]
	\centering
	\includegraphics[width=0.95\linewidth]{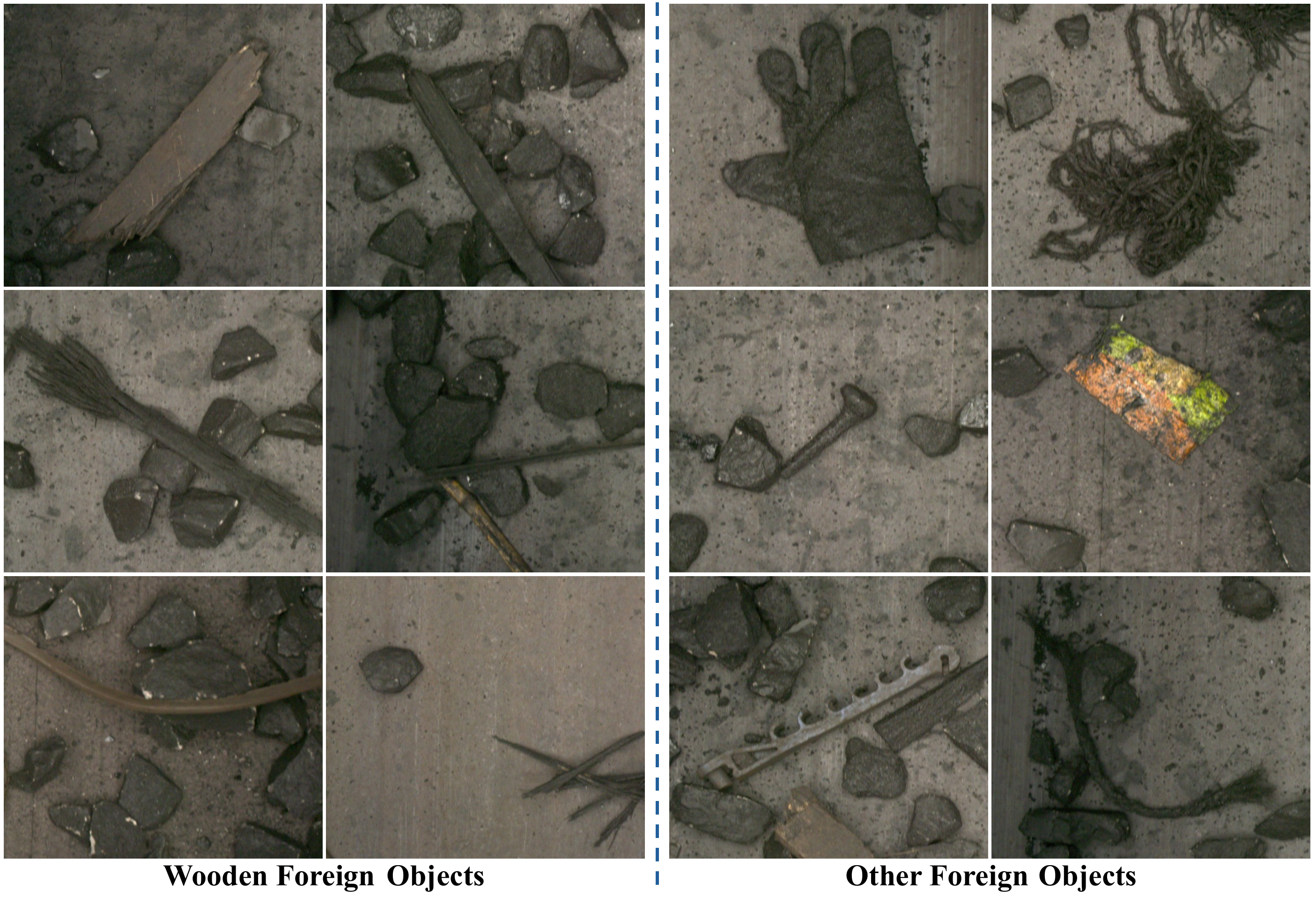}
	\caption{\textbf{Example foreign objects in the CoalAD dataset.} The left six samples show wooden objects, while the right six depict other types (e.g., nets and metal parts). Notably, even within the same material category (wood), the appearances vary substantially due to compression, staining, and surface contamination.}
	\label{fig:Coal_Objects}
\end{figure}

The training set contains 2,490 normal images. The test set includes 811 normal samples and 943 anomalous samples. Among the anomalous samples, clearly identifiable wooden foreign objects account for the largest portion (799 images), which is consistent with their relatively high occurrence frequency in practice (e.g., wooden support structures commonly used in mining operations). The remaining anomalous samples cover various other types such as nets, metal braces/rods, gloves, and plastic bags. Owing to factors such as compression and staining, wooden foreign objects themselves exhibit diverse visual appearances. Fig.~\ref{fig:Coal_Objects} illustrates representative examples of wooden and other foreign objects in CoalAD.

\section{Method}
\subsection{Overview}
Our goal is to learn the normal visual patterns of conveyor-belt coal-stream scenes using only normal training data.
At inference time, given an input image $X \in \mathbb{R}^{H \times W \times 3}$, our method produces (i) an image-level anomaly score $s \in \mathbb{R}$ to determine whether $X$ contains foreign objects, and (ii) a pixel-level anomaly heatmap $M \in \mathbb{R}^{H \times W}$ to localize anomalous regions, enabling detection of unseen foreign objects with pixel-level localization.

Existing methods achieve strong performance in structured industrial scenarios such as MVTec AD largely because they rely on an implicit assumption: that in normal samples, local appearances and their correspondences to global context are relatively stable. Under this premise, reconstruction/generation/distillation-based approaches typically learn local predictability and representation consistency under normal conditions, and treat significant local deviations from such consistency as anomaly cues. However, in coal-stream conveyor scenarios, the piling morphology, particle scale, and occlusion patterns of coal and gangue vary continuously, making local textures and their contextual relations inherently unstable even for normal data. This weakens the reliability of the above ``consistency-relation'' signals and may cause the model to become overly sensitive to local texture perturbations, leading to degraded performance when foreign objects are texturally similar to the background.

In contrast, PatchCore-style methods based on local patch memory matching mainly rely on whether a local patch falls within the neighborhood of normal local features. 
This can be advantageous in coal-stream scenes where ``consistency-relation'' cues degrade: storing diverse normal patch exemplars provides a more direct and comprehensive characterization of normal local textures, which can sometimes be more robust than relation-modeling approaches under highly variable normal conditions. 
Nevertheless, relying solely on local similarity still lacks higher-level semantic constraints: anomalies that primarily disrupt global semantic consistency may not exhibit salient local texture deviations and can thus be missed. 
Conversely, when local appearances are heavily corrupted or ambiguous (e.g., due to occlusion or coal-dust contamination), local matching may become noisy and trigger false alarms.

To address these challenges, we propose a three-branch anomaly detection framework that fuses complementary anomaly cues from an object-level branch, a semantic attribution branch, and a texture branch. Specifically, the three branches capture complementary evidence from \textbf{object semantic structure, the distribution of global semantic contributions, and fine-grained texture patterns}. These cue-specific deviation signals are fused at inference time to produce robust image-level anomaly scores and accurate pixel-level localization.

\subsection{Object-level Branch}
\label{ssec:object_branch}

\begin{figure*}[t]
	\centering
	\includegraphics[width=1\linewidth]{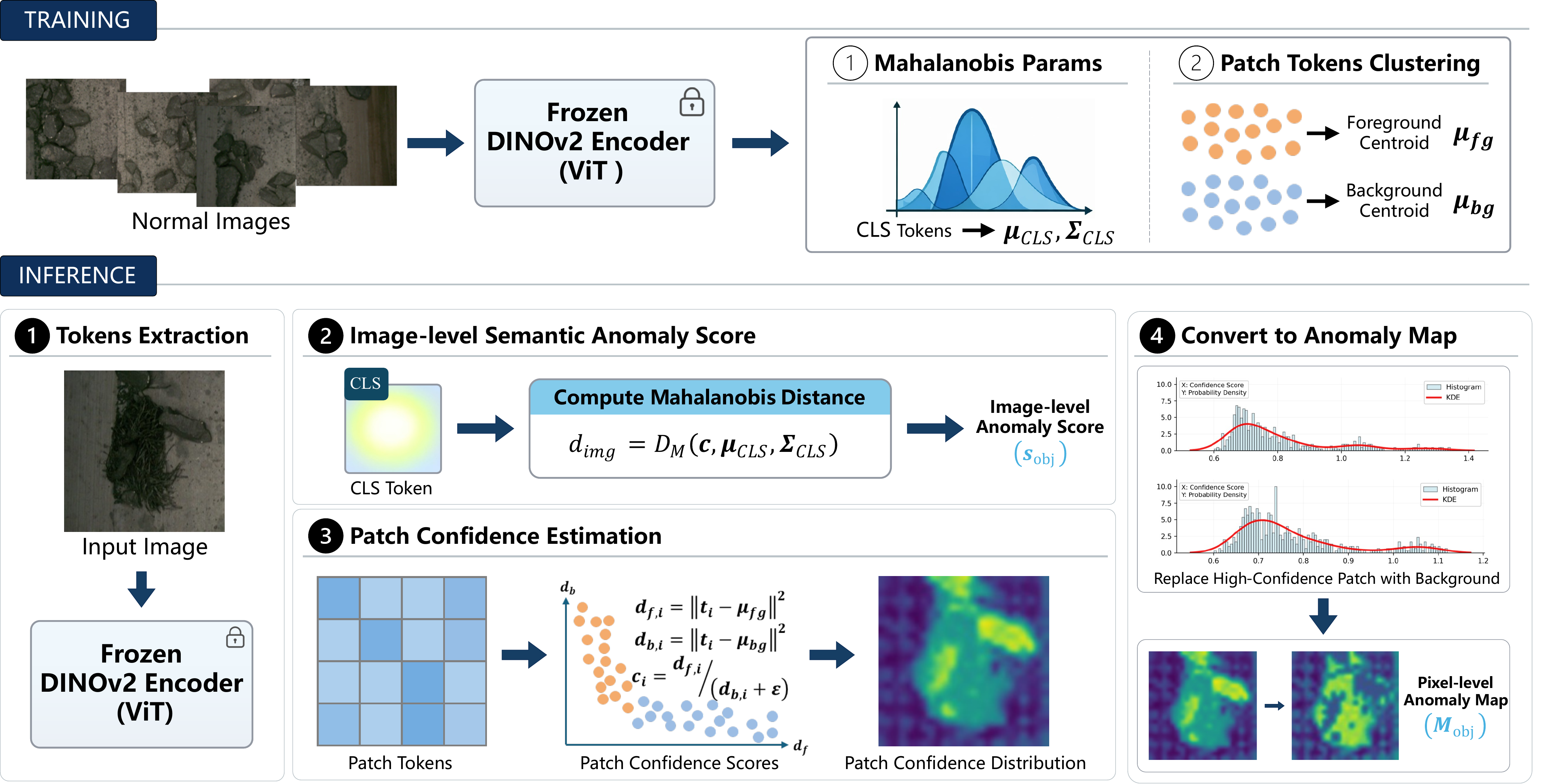}
	\caption{\textbf{Pipeline of the object-level branch.} From normal data, frozen DINOv2 features are used to learn a global CLS distribution and two semantic anchors for the normal foreground (coal/gangue) and background (conveyor belt). At test time, global semantic deviation yields an image-level anomaly score, and per-patch deviation to these normal anchors yields pixel-level anomaly scores for localization.}
	\label{fig:MACLU}
\end{figure*}

\subsubsection{Branch Objective and Design Rationale}
The overall pipeline of this branch is illustrated in Fig.~\ref{fig:MACLU}. Our goal is to learn the \emph{semantic patterns of objects that typically appear in normal coal-stream scenes} using only normal images during training, and to determine anomalies at inference time according to the degree to which an input image deviates from such semantic patterns.

This goal implies a dual modeling approach:
\begin{itemize}
	\item \textbf{Image-level Modeling:} Statistically model the global semantic representations of all normal samples. An image-level anomaly score is derived from the deviation of a test sample from this normal semantic distribution.
	\item \textbf{Object-level Modeling:} Model the semantic types/composition of objects that commonly occur in normal images. This enables the identification and pixel-level localization of regions that do not conform to normal semantic types.
\end{itemize}
Therefore, this branch critically relies on a feature extractor that provides strong object-level semantics, and whose local features are well-aligned with semantic content (i.e., semantically meaningful and spatially localized).

\subsubsection{Feature Extractor: Frozen DINOv2-ViT}
To meet the above requirement, we adopt a frozen DINOv2-ViT as the feature extractor. DINOv2 learns general-purpose visual representations via large-scale self-supervised pretraining and exhibits strong cross-task transferability, supporting both image-level and pixel-level downstream tasks without task-specific fine-tuning~\cite{DINOv2}. Moreover, prior studies have shown that self-supervised ViTs (e.g., DINO) can exhibit emergent object-centric cues in attention maps and token embeddings, which have been exploited for unsupervised object discovery/localization and segmentation~\cite{DINO-Emerging,LOST}. Recent works further demonstrate that DINOv2 features serve as strong dense visual descriptors and backbones for correspondence-related tasks (e.g., dense matching and semantic correspondence)~\cite{RoMa,SDDINO}.

Given an input image \(x \in \mathbb{R}^{H \times W \times 3}\), the frozen encoder \(E_{\text{DINOv2}}\) extracts token representations:
\begin{equation}
	[\mathbf{c}, \mathbf{P}] = E_{\text{DINOv2}}(x),
\end{equation}
where \(\mathbf{c} \in \mathbb{R}^{D}\) is the global \texttt{[CLS]} token and \(\mathbf{P} = \{\mathbf{p}_i\}_{i=1}^{N}\) is the set of \(N\) patch tokens, each of dimension \(D\).

\subsubsection{Training Phase: Building Normal References}
Using only normal training images, we construct two types of \emph{normal references} from normal training samples: a statistical distribution for image-level semantics, and foreground/background prototypes for object composition.

\textbf{1. Global Semantic Distribution (CLS Statistics).}
We collect the CLS tokens from all \(K\) normal training samples, \(\mathcal{C} = \{\mathbf{c}_k\}_{k=1}^{K}\), and estimate their mean and covariance:
\begin{equation}
	\boldsymbol{\mu}_{\text{CLS}}=\frac{1}{K}\sum_{k=1}^{K}\mathbf{c}_k,\qquad
	\boldsymbol{\Sigma}_{\text{CLS}}=\frac{1}{K-1}\sum_{k=1}^{K}(\mathbf{c}_k-\boldsymbol{\mu}_{\text{CLS}})(\mathbf{c}_k-\boldsymbol{\mu}_{\text{CLS}})^{\top}.
\end{equation}
These statistics define the expected distribution of global scene semantics and are used to measure deviation at test time.

\textbf{2. Foreground/Background Prototypes (Patch Clustering).}
All patch tokens from normal training images are aggregated into a set \(\mathcal{P}\). In the coal-stream conveyor scenario, normal samples mainly contain two semantic categories: coal/coal-gangue (foreground) and the conveyor belt (background). Accordingly, we perform \textbf{two-cluster KMeans clustering} (\(K=2\)) on \(\mathcal{P}\), obtaining cluster centers \(\boldsymbol{\mu}_0\) and \(\boldsymbol{\mu}_1\).

We semantically align these clusters: the cluster with fewer patches is designated as the \textbf{foreground} prototype \(\boldsymbol{\mu}_{fg}\) (coal/gangue), and the larger cluster as the \textbf{background} prototype \(\boldsymbol{\mu}_{bg}\) (conveyor belt):
\begin{equation}
	(\boldsymbol{\mu}_{fg},\boldsymbol{\mu}_{bg})=
	\begin{cases}
		(\boldsymbol{\mu}_0,\boldsymbol{\mu}_1), & |\mathcal{P}_0|<|\mathcal{P}_1|,\\
		(\boldsymbol{\mu}_1,\boldsymbol{\mu}_0), & |\mathcal{P}_0|\ge|\mathcal{P}_1|,
	\end{cases}
\end{equation}
where \(|\mathcal{P}_j|\) denotes the size of cluster \(j\).

\subsubsection{Inference Phase: Anomaly Scoring and Localization}
For a test image with extracted tokens \([\mathbf{c}, \mathbf{P}]\), we compute the image-level score and pixel-level heatmap as follows.

\textbf{Image-Level Anomaly Score.}
The deviation of the test image's global semantics from the learned normal distribution is quantified by the Mahalanobis distance:
\begin{equation}
	s_{\text{obj}} \triangleq \sqrt{(\mathbf{c}-\boldsymbol{\mu}_{\text{CLS}})^{\top}\boldsymbol{\Sigma}_{\text{CLS}}^{-1}(\mathbf{c}-\boldsymbol{\mu}_{\text{CLS}})}.
\end{equation}
Intuitively, when a foreign object appears, the global semantic representation is more likely to deviate from the semantic distribution of normal coal scenes, leading to a higher score.

\textbf{Pixel-Level Anomaly Heatmap.}
The heatmap is generated through a three-step process:
\begin{enumerate}
	\item \textbf{Patch Response Calculation:} For each patch token \(\mathbf{p}_i\), compute its Euclidean distances to the learned prototypes:
	\begin{equation}
		d_{fg}(i) = \|\mathbf{p}_i-\boldsymbol{\mu}_{fg}\|_2, \quad d_{bg}(i) = \|\mathbf{p}_i-\boldsymbol{\mu}_{bg}\|_2.
	\end{equation}
	A foreground-biased score \(r_i\) is defined as:
	\begin{equation}
		r_i = \frac{d_{bg}(i)}{d_{fg}(i) + \varepsilon}.
	\end{equation}
	A higher \(r_i\) indicates a stronger resemblance to the foreground.
	
	\item \textbf{Mean Replacement for Anomaly Highlighting:} In practice, patches corresponding to true coal/gangue foreground often yield very high \(r_i\) values. To suppress these strong foreground responses and to highlight regions that belong to \textit{neither} the learned foreground nor background distributions (i.e., potential foreign objects), we apply a mean replacement strategy. For a threshold \(\tau\):
	\begin{equation}
		\tilde r_i = \begin{cases}
			r_i, & r_i \le \tau,\\
			\mathbb{E}[r \mid bg], & r_i > \tau,
		\end{cases}
		\qquad \text{where } \mathbb{E}[r \mid bg] = \frac{1}{|\mathcal{I}_{bg}|} \sum_{j \in \mathcal{I}_{bg}} r_j.
	\end{equation}
	Here, \(\mathcal{I}_{bg}\) contains indices of patches predicted as background by the KMeans model. Thus, \(\tilde r_i\) becomes a patch-level anomaly score where a larger value indicates a higher likelihood of being an anomaly.
	
	\item \textbf{Pixel-wise Score Map:} The scores $\{\tilde r_i\}$ are reshaped into a patch grid $\tilde R \in \mathbb{R}^{H_p \times W_p}$ and upsampled to the original image resolution. We then apply Gaussian smoothing to reduce block artifacts, yielding a pixel-wise anomaly score map:
	\begin{equation}
		M_{\text{obj}} = \text{Gauss}\!\left(\text{Upsample}(\tilde R)\right) \in \mathbb{R}^{H \times W}.
	\end{equation}
	Here, $\text{Upsample}(\cdot)$ denotes interpolation-based upsampling, and $\text{Gauss}(\cdot)$ denotes Gaussian smoothing.
	
\end{enumerate}

\subsection{Semantic Attribution Branch}
\label{ssec:semantic_branch}

\begin{figure*}[t]
	\centering
	\includegraphics[width=1\linewidth]{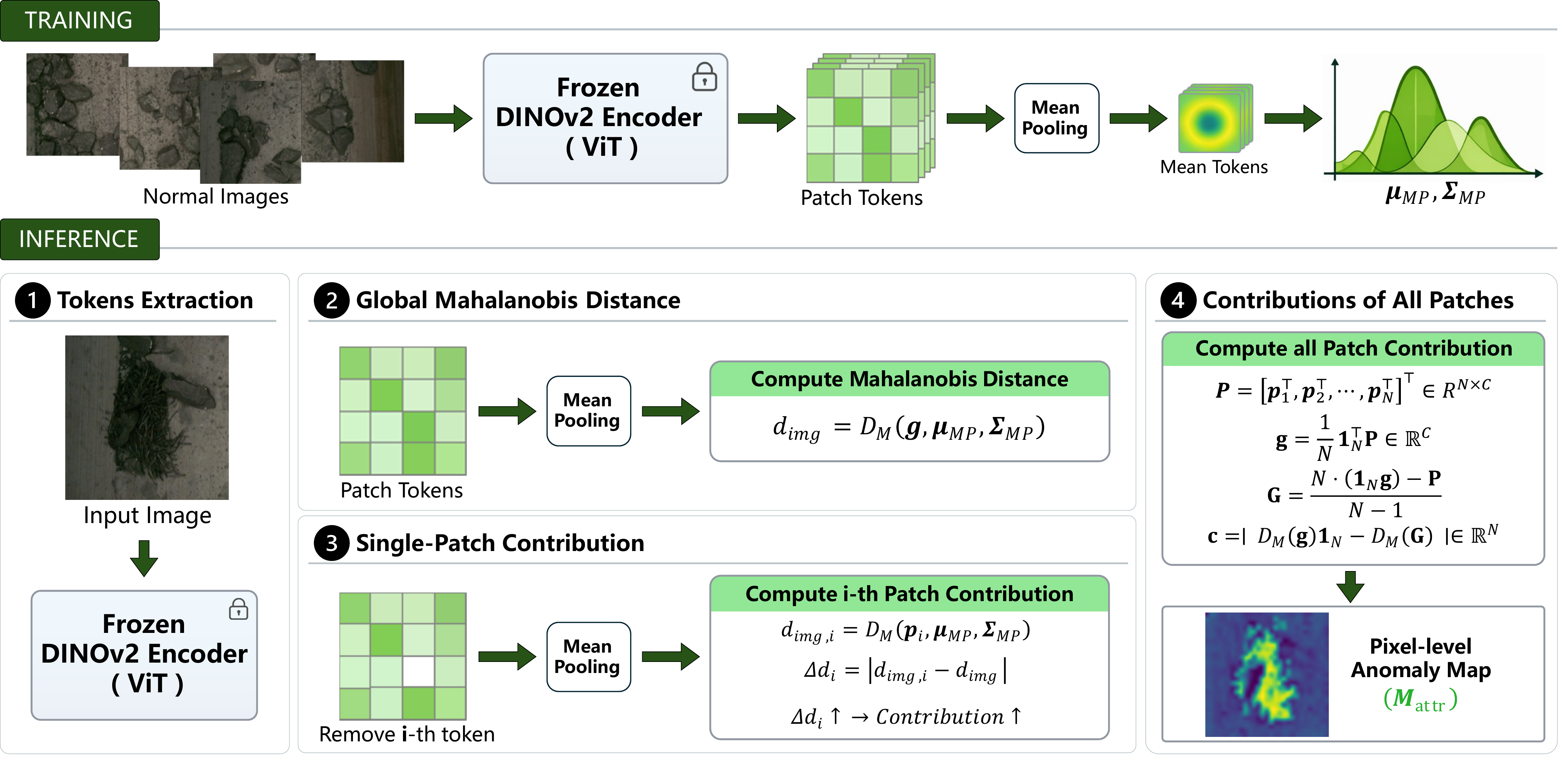}
	\caption{\textbf{Pipeline of the semantic attribution branch.} The branch constructs a normal semantic baseline from mean-pooled patch features. At inference, it attributes the global semantic deviation of a test image to individual patches via closed-form ablation, producing an attribution-based anomaly score map for localization.}
	\label{fig:Ma2Patch}
\end{figure*}

\subsubsection{Branch Objective and Design Rationale}
The overall pipeline of this branch is illustrated in Fig.~\ref{fig:Ma2Patch}. Our goal is to build a top-down semantic attribution mechanism that quantifies each patch's contribution to the global semantic deviation of an image. The core premise is that patches which contribute more to the deviation from the normal semantic baseline have a higher likelihood of being anomalous.

Similar to the object-level branch, this design critically relies on a feature extractor with strong semantic representation capacity and a reliable correspondence between local features and high-level semantics. Therefore, we also adopt a frozen DINOv2-ViT as the feature extractor.

The key distinction between the two branches lies in their focus: while the object-level branch emphasizes learning the typical semantic composition (foreground vs. background) of normal scenes, this branch focuses on explaining \emph{where} an anomalous global semantic signal originates. This attribution-driven perspective provides a complementary cue for localizing unknown foreign objects.

\subsubsection{Feature Representation with DINOv2-ViT}
We keep the encoder \(E_{\text{DINOv2}}\) frozen during both training and inference. Given an input image \(x \in \mathbb{R}^{H \times W \times 3}\), it extracts token representations:
\begin{equation}
	[\mathbf{c}, \mathbf{P}] = E_{\text{DINOv2}}(x),
\end{equation}
where \(\mathbf{c} \in \mathbb{R}^{D}\) is the \texttt{[CLS]} token and \(\mathbf{P} = \{\mathbf{p}_i\}_{i=1}^{N}\) is the set of \(N\) patch tokens.

In this branch, we do not use the \texttt{[CLS]} token directly. Instead, we construct a global semantic vector \(\mathbf{g}\) by mean-pooling all patch tokens:
\begin{equation}
	\mathbf{g} \triangleq \frac{1}{N}\sum_{i=1}^{N}\mathbf{p}_i \in \mathbb{R}^{D}.
\end{equation}
This mean-pooled vector directly aggregates patch-level information, forming a representation that is intrinsically tied to local patches and thus suitable for subsequent attribution analysis.

\subsubsection{Training: Establishing the Normal Semantic Baseline}
Using only normal training images, we statistically model the distribution of their global semantic vectors ($\mathbf{g}$) to establish a reference for measuring deviation.

\textbf{1. Aggregate Normal Representations.}
We collect the mean-pooled global semantic vectors from all $K$ normal training samples: $\mathcal{G}=\{\mathbf{g}_k\}_{k=1}^{K}$.

\textbf{2. Statistical Modeling of the Distribution.}
We estimate the mean $\boldsymbol{\mu}_{\text{MP}}$ and covariance $\boldsymbol{\Sigma}_{\text{MP}}$ of this set:
\begin{equation}
	\boldsymbol{\mu}_{\text{MP}}=\frac{1}{K}\sum_{k=1}^{K}\mathbf{g}_k,\qquad
	\boldsymbol{\Sigma}_{\text{MP}}=\frac{1}{K-1}\sum_{k=1}^{K}(\mathbf{g}_k-\boldsymbol{\mu}_{\text{MP}})(\mathbf{g}_k-\boldsymbol{\mu}_{\text{MP}})^{\top}.
\end{equation}

The pair $(\boldsymbol{\mu}_{\text{MP}},\boldsymbol{\Sigma}_{\text{MP}})$ defines the Mahalanobis distance metric for measuring semantic deviation and serves as the baseline for attribution.

\subsubsection{Inference: Attribution-based Anomaly Localization}
For a test image with patch tokens \(\mathbf{P}\) and its global vector \(\mathbf{g}\), we compute an image-level anomaly score and a pixel-level heatmap through attribution analysis.

\textbf{Image-Level Anomaly Score.}
The deviation of the test image’s global semantics from the normal baseline is quantified by the Mahalanobis distance:
\begin{equation}
	s_{\text{attr}} \triangleq \sqrt{(\mathbf{g} - \boldsymbol{\mu}_{\text{MP}})^{\top} \boldsymbol{\Sigma}_{\text{MP}}^{-1} (\mathbf{g} - \boldsymbol{\mu}_{\text{MP}})}.
\end{equation}

\textbf{Pixel-Level Anomaly Heatmap via Attribution.}
To localize the source of the global deviation \(s_{\text{attr}}\), we estimate each patch's contribution through a closed-form, ablation-style derivation. This process consists of three main steps:
\begin{enumerate}
	\item \textbf{Matrix Formulation.} Stack the patch tokens into a matrix \(\mathbf{P}_{\text{mat}} \in \mathbb{R}^{N \times D}\) and note that \(\mathbf{g} = \frac{1}{N}\mathbf{1}_N^{\top}\mathbf{P}_{\text{mat}}\).
	
	\item \textbf{Closed-Form Patch Ablation.} The global vector with the \(i\)-th patch ablated is \(\mathbf{g}_{-i} = (N\mathbf{g} - \mathbf{p}_i) / (N-1)\). We compute all such ablated vectors efficiently in matrix form:
	\begin{equation}
		\mathbf{G} = \frac{N \cdot (\mathbf{1}_N \mathbf{g}) - \mathbf{P}_{\text{mat}}}{N-1} \in \mathbb{R}^{N \times D},
	\end{equation}
	where the \(i\)-th row of \(\mathbf{G}\) equals \(\mathbf{g}_{-i}\).
	
	\item \textbf{Contribution Scoring.} Let \(D_M(\cdot)\) denote the Mahalanobis distance defined by \((\boldsymbol{\mu}_{\text{MP}}, \boldsymbol{\Sigma}_{\text{MP}})\). Compute the full distance \(d_{\text{full}} = D_M(\mathbf{g})\) and the vector of ablated distances \(\mathbf{d} = D_M(\mathbf{G}) \in \mathbb{R}^{N}\) (applied row-wise). The contribution \(c_i\) of the \(i\)-th patch is defined as the absolute change in distance induced by its removal:
	\begin{equation}
		c_i \triangleq \left| d_{\text{full}} - \mathbf{d}_i \right|, \qquad i = 1, \dots, N.
	\end{equation}
	A larger \(c_i\) signifies that the patch is more responsible for the overall semantic deviation.
\end{enumerate}

Finally, the patch contributions $\{c_i\}_{i=1}^{N}$ are reshaped into a patch grid $C \in \mathbb{R}^{H_p \times W_p}$.
This grid is then upsampled to the original image resolution and Gaussian-smoothed to generate the final pixel-level anomaly map:
\begin{equation}
	M_{\text{attr}} = \text{Gauss}\!\left(\text{Upsample}(C)\right) \in \mathbb{R}^{H \times W}.
\end{equation}

\subsection{Texture Branch}
\label{ssec:texture_branch}

\begin{figure}[t]
	\centering
	\includegraphics[width=1\linewidth]{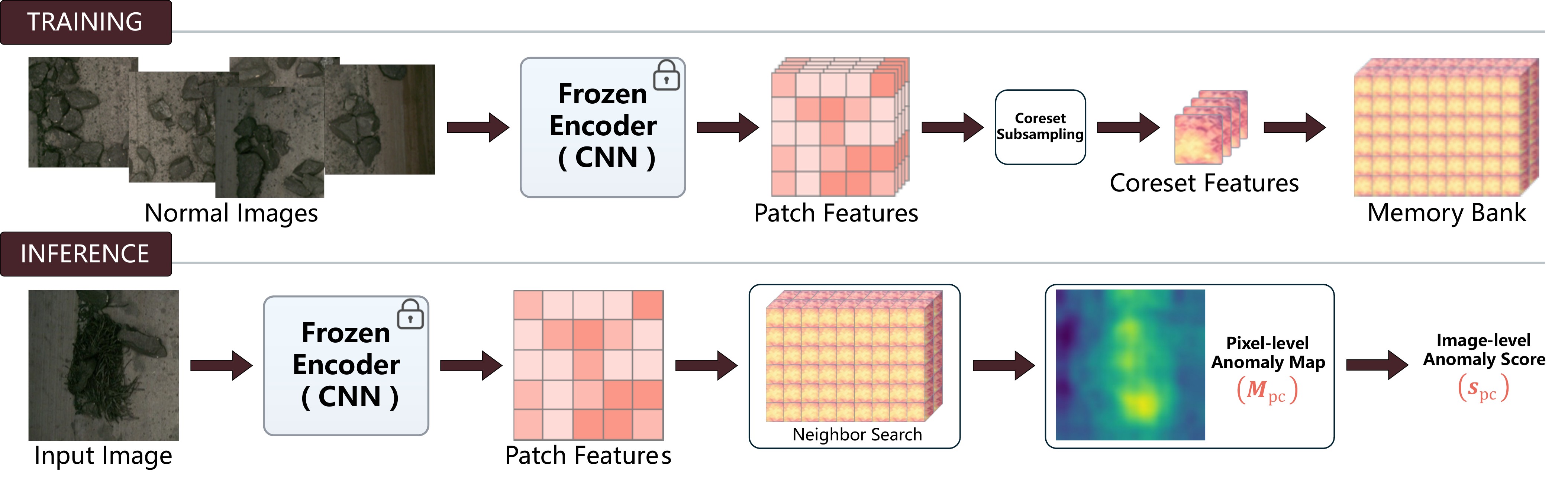}
	\caption{\textbf{Pipeline of the texture branch.} Based on the PatchCore framework, we build a memory bank of normal local features that capture fine-grained texture/structure patterns. At inference, nearest-neighbor distances to the normal feature bank provide patch-wise anomaly scores, which are aggregated into an image-level score and a pixel-level anomaly score map.}
	\label{fig:patchcore}
\end{figure}

\subsubsection{Branch Objective and Design Rationale}
The architecture of this branch is illustrated in Fig.~\ref{fig:patchcore}. It serves as a complementary branch in our framework, aiming to capture anomalies manifested in fine-grained local textures and structures. Concretely, we adopt PatchCore as this texture/structure branch: it constructs a memory bank of normal local patch representations from multi-scale CNN features during training. At inference time, it measures the deviation of test patches from this normal reference via nearest-neighbor matching, producing a pixel-level anomaly map $M_{\text{pc}}$ and an image-level anomaly score $s_{\text{pc}}$.

The complementarity of this branch lies in two aspects:
\begin{itemize}
	\item \textbf{Focus on Local Patterns:} Unlike Vision Transformers which excel at capturing high-level semantics, CNNs are inherently more sensitive to local textures and fine-grained spatial patterns due to their convolutional inductive bias. This makes them adept at detecting subtle anomalous details that may not significantly alter global scene semantics.
	\item \textbf{Non-parametric Matching:} The memory bank provides an exemplar-based, non-parametric reference that covers diverse normal patterns and complements the parametric semantic modeling in the first two branches.
\end{itemize}

To capture the fine-grained local patterns that are crucial for texture anomaly detection, we employ a frozen CNN encoder (pretrained on ImageNet). No parameters are updated during training; the encoder is used solely for feature extraction.

\subsubsection{Training Phase: Memory Bank Construction}
Using normal training images $\{x_n\}_{n=1}^{K}$, we build a compact memory bank of normal patch representations through the following steps:
\begin{enumerate}
	\item \textbf{Multi-scale Feature Extraction:} For each image, we extract intermediate feature maps $\{F^{(l)}\}$ from multiple layers of $E_{\text{CNN}}$.
	\item \textbf{Patch Representation Generation:} These multi-scale features are spatially aligned (e.g., via resizing) and aggregated (e.g., concatenation) to form a dense set of local patch representations $\mathcal{P}=\{p_i\}$ over the entire training set.
	\item \textbf{Coreset Subsampling:} To improve efficiency, we apply coreset subsampling to $\mathcal{P}$ and retain a representative subset as the memory bank:
	\begin{equation}
		\mathcal{M}=\mathrm{Coreset}(\mathcal{P}), \qquad |\mathcal{M}|\approx \lfloor f \cdot |\mathcal{P}| \rfloor,
	\end{equation}
	where $f$ is the coreset ratio.
\end{enumerate}
This process involves no parameter learning; it solely constructs the non-parametric reference set $\mathcal{M}$.

\subsubsection{Inference Phase: Nearest-Neighbor Scoring \& Localization}
For a test image \(x\), we extract its patch representations \(\{p_i^{\text{test}}\}\) using the same feature extractor \(E_{\text{CNN}}\).

\textbf{Pixel-Level Anomaly Map.}
For each test patch \(p_i^{\text{test}}\), we compute its distance to the nearest neighbor in the memory bank \(\mathcal{M}\):
\begin{equation}
	d_i = \min_{m \in \mathcal{M}} \| p_i^{\text{test}} - m \|_2.
\end{equation}
The set \(\{d_i\}\) is spatially reshaped into a low-resolution anomaly map, which is then upsampled to the input image size and smoothed with a Gaussian kernel for spatial consistency:
\begin{equation}
	M_{\text{pc}} = \mathrm{Gauss}\left( \mathrm{Upsample}\left( \mathrm{Reshape}(\{d_i\}) \right) \right).
\end{equation}

\textbf{Image-Level Anomaly Score.}
The image-level score is derived from the most anomalous patch response, re-weighted to reduce spurious isolated responses:
\begin{equation}
	s_{\text{pc}} = w \cdot \max_i d_i.
\end{equation}
The weight $w$ is computed based on how ``rare/sparse'' the nearest memory-bank neighbor $m^{*}$ is with respect to its neighbors in $\mathcal{M}$, following the re-weighting scheme in PatchCore.

\subsection{Fused Inference}
\label{ssec:fused_inference}

\begin{figure}[t]
	\centering
	\includegraphics[width=1\linewidth]{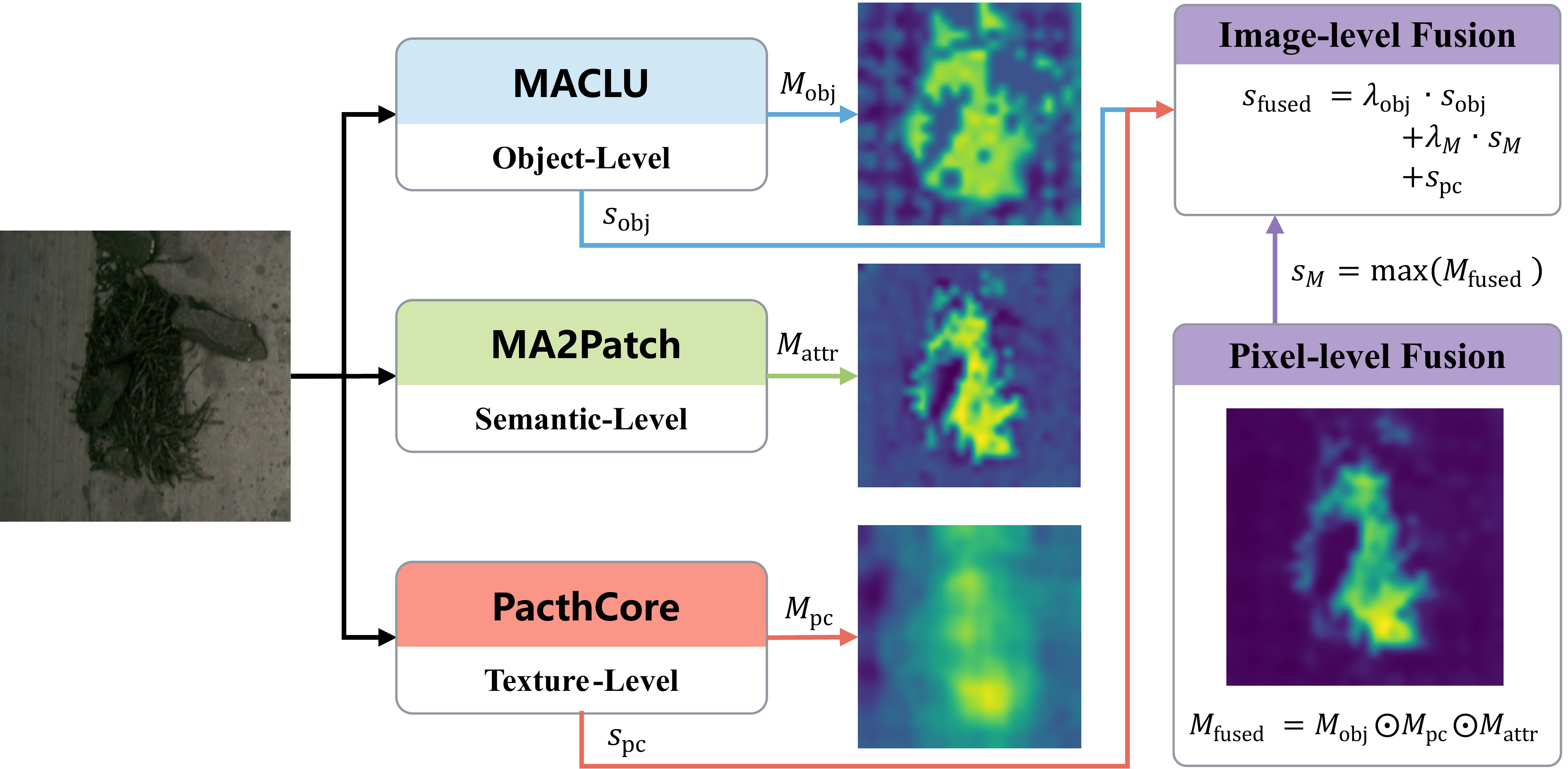}
	\caption{\textbf{Pipeline of the multi-branch fusion strategy.} The anomaly maps and scores from the three branches are aligned and fused via Hadamard product (pixel-level) and weighted combination (image-level) to produce the final output.}
	\label{fig:fusion_inference}
\end{figure}

As illustrated in Fig.~\ref{fig:fusion_inference}, our framework runs the three complementary branches in parallel during inference on the same input image \(x\):
\begin{itemize}
	\item The \textbf{object-level branch (\textbf{MACLU}\footnote{\textbf{MA}halanobis + \textbf{CLU}stering, denoting its two core operations.})}: Provides evidence based on deviations from learned object composition.
	\item The \textbf{semantic attribution branch (\textbf{MA2Patch}\footnote{\textbf{MA}halanobis-\textbf{to-Patch} attribution, indicating its mapping of global deviation to patch-level contributions.})}: Attributes global semantic deviations to local patches.
	\item The \textbf{texture branch (based on \textbf{PatchCore}~\cite{PatchCore})}: Detects anomalies via fine-grained local texture/structure matching.
\end{itemize}
The outputs from these branches are fused at both the pixel level and the image level to produce the final anomaly localization heatmap and detection score.

\subsubsection{Pixel-Level Fusion: Consensus Gating via Hadamard Product}
Given a test image \(x\), each branch produces pixel-level evidence maps:
\begin{itemize}
	\item \(M_{\text{obj}}\): Foreground/background deviation map from \textbf{MACLU}.
	\item \(M_{\text{attr}}\): Semantic attribution map from \textbf{MA2Patch}.
	\item \(M_{\text{pc}}\): Nearest-neighbor distance map from the texture branch.
\end{itemize}
To fuse these maps, we first ensure they are compatible in scale and resolution.

\textbf{1. Normalization and Scale Alignment.}
The maps \(M_{\text{obj}}\) and \(M_{\text{pc}}\) are min-max normalized to the range \([0, 1]\):
\begin{equation}
	\tilde{M} = \frac{M - \min(M)}{\max(M) - \min(M) + \varepsilon}.
\end{equation}
The attribution map \(M_{\text{attr}}\) is kept unnormalized to preserve its original response magnitudes. All maps are then resized via interpolation to a common spatial resolution \((H, W)\), denoted as \(\hat{M}_{\text{obj}}\), \(\hat{M}_{\text{attr}}\), and \(\hat{M}_{\text{pc}}\). This normalization/alignment step makes heterogeneous cues comparable before fusion, enabling a more effective integration of multi-branch evidence~\cite{MAVD}.

\textbf{2. Consensus Gating Fusion.}
The final pixel-level anomaly heatmap is obtained through element-wise (Hadamard) multiplication:
\begin{equation}
	M_{\text{fused}} = \hat{M}_{\text{obj}} \odot \hat{M}_{\text{pc}} \odot \hat{M}_{\text{attr}}.
	\label{eq:pixel_fusion}
\end{equation}
This operation implements a \emph{consensus gating} mechanism:
\begin{itemize}
	\item The normalized maps \(\hat{M}_{\text{obj}}\) and \(\hat{M}_{\text{pc}}\) act as soft gates in \([0,1]\).
	\item They adaptively modulate the attribution responses in \(\hat{M}_{\text{attr}}\): regions supported by both object-level and texture-level evidence are amplified, while unsupported or noisy attribution responses are suppressed.
\end{itemize}
We treat \(\hat{M}_{\text{attr}}\) as the fusion backbone because it directly reflects global semantic anomalies, but it may lack fine-grained detail. The gating from \(\hat{M}_{\text{obj}}\) and \(\hat{M}_{\text{pc}}\) complements this by injecting localized object-composition and texture cues. Finally, a lightweight Gaussian smoothing is applied to \(M_{\text{fused}}\) to suppress isolated noise and improve spatial coherence.

\subsubsection{Image-Level Fusion: Weighted Aggregation with Peak Compensation}
At the image level, we aggregate the anomaly scores from each branch into a final detection score.

\textbf{1. Base Score Fusion.}
We combine the image-level scores from the texture branch (\(s_{\text{pc}}\)) and the object-level branch (\(s_{\text{obj}}\)) with a scaling weight \(\lambda_{\text{obj}}\) to balance their numerical scales:
\begin{equation}
	s_{\text{base}} = s_{\text{pc}} + \lambda_{\text{obj}} \cdot s_{\text{obj}}.
	\label{eq:image_fusion_base}
\end{equation}

\textbf{2. Peak-Value Compensation.}
To explicitly propagate strong local anomalies identified in the pixel-level fusion to the global decision, we use the peak value of the fused heatmap \(M_{\text{fused}}\) as a compensation term. This term is weighted by \(\lambda_M\):
\begin{equation}
	s_{\text{fused}} = s_{\text{base}} + \lambda_M \cdot \max\big(M_{\text{fused}}\big).
	\label{eq:image_fusion_final}
\end{equation}
The weight \(\lambda_M\) controls the influence of this spatially aware cue on the final image-level score.

\section{Experiments}
\subsection{Dataset and Evaluation Metrics}
\label{sec:exp_dataset_metrics}

\subsubsection{Dataset: CoalAD}
We conduct experiments on our constructed \textbf{CoalAD} benchmark for unsupervised foreign-object inspection in coal streams. Its detailed characteristics and construction process are described in Section~\ref{sec:dataset}. Following the benchmark protocol, models are trained exclusively on 2,490 normal samples (coal and gangue). The test set contains 811 normal and 943 anomalous images with foreign objects, requiring models to perform both image-level detection and pixel-level localization.

\subsubsection{Evaluation Metrics}
To comprehensively evaluate model performance, we adopt three standard metrics widely used in industrial anomaly detection:
\begin{itemize}
	\item \textbf{Image-level ROC-AUC:} Evaluates the model's ability to discriminate between normal and anomalous images using the image-level anomaly score.
	\item \textbf{Pixel-level ROC-AUC:} Assesses the quality of the pixel-wise anomaly map by measuring per-pixel discrimination against the binary ground-truth mask.
	\item \textbf{PRO-AUC (Per-Region Overlap):} Quantifies localization accuracy by computing the overlap between predicted anomalous regions and each connected component in the ground truth across varying thresholds. We report the area under the PRO curve at false positive rates up to 0.3.
\end{itemize}
Image-level AUC reflects the system's primary detection reliability, while pixel-level AUC and PRO-AUC are critical for providing spatial guidance for downstream operations such as robotic removal.

\subsection{Implementation Details}
\label{sec:exp_implementation}

All experiments are conducted on a workstation with an Intel i7-11700T CPU, 64 GB RAM, and an NVIDIA GeForce RTX 3090 Ti GPU. The code is implemented in Python 3.10 with PyTorch and CUDA. For reproducibility, we fix the random seed to 22.

Our three-branch framework uses frozen, pre-trained feature extractors without task-specific fine-tuning. The configuration for each branch is as follows:
\begin{itemize}
	\item \textbf{Object-level Branch (\textbf{MACLU}):} Uses a frozen DINOv2 ViT-Small backbone with an input resolution of $350 \times 350$.
	\item \textbf{Semantic Attribution Branch (\textbf{MA2Patch}):} Uses a frozen DINOv2 ViT-Base backbone with an input resolution of $406 \times 406$.
	\item \textbf{Texture Branch (\textbf{PatchCore}):} Uses a frozen Wide-ResNet50-2 backbone with an input resolution of $238 \times 238$.
\end{itemize}

Key hyperparameters are set as follows:
\begin{itemize}
	\item \textbf{MACLU:} KMeans clustering with $K=2$; strong-foreground suppression threshold $\tau = 1.14$ for patch replacement.
	\item \textbf{PatchCore:} Coreset subsampling ratio $f_{\text{coreset}} = 0.01$; number of nearest neighbors $k = 3$ for score re-weighting; sparse projection parameter $\epsilon_{\text{coreset}} = 0.90$.
\end{itemize}
For each experimental split, the PatchCore memory bank and the statistical models in MACLU/MA2Patch are built using only the corresponding normal training samples and evaluated on the held-out test set.

\subsection{Comparison with Representative Baselines}
\label{sec:exp_sota}

\begin{table*}[t]
	\centering
	\small
	\setlength{\tabcolsep}{5.0pt}
	\renewcommand{\arraystretch}{1.18}
	
	\resizebox{\textwidth}{!}{%
		\begin{tabular}{lccccccccc}
			\toprule
			& PatchCore & SimpleNet & DRAEM & ISSTAD & EfficientAD & RD4AD & RD++ & CKAAD & \textbf{Ours} \\
			\cmidrule(lr){2-2}\cmidrule(lr){3-3}\cmidrule(lr){4-4}\cmidrule(lr){5-5}\cmidrule(lr){6-6}\cmidrule(lr){7-7}\cmidrule(lr){8-8}\cmidrule(lr){9-9}\cmidrule(lr){10-10}
			\multicolumn{1}{c}{} &
			\multicolumn{1}{c}{\scriptsize CVPR'22} &
			\multicolumn{1}{c}{\scriptsize CVPR'23} &
			\multicolumn{1}{c}{\scriptsize CVPR'21} &
			\multicolumn{1}{c}{\scriptsize EAAI'25} &
			\multicolumn{1}{c}{\scriptsize WACV'24} &
			\multicolumn{1}{c}{\scriptsize CVPR'22} &
			\multicolumn{1}{c}{\scriptsize CVPR'23} &
			\multicolumn{1}{c}{\scriptsize AAAI'25} &
			\multicolumn{1}{c}{\scriptsize --} \\
			\midrule
			Image-AUC (\%) $\uparrow$ & \underline{94.47} & 92.18 & 65.06 & 80.27 & 87.93 & 85.10 & 89.24 & 85.62 & \cellcolor{gray!12}\textbf{98.83} \\
			Pixel-AUC (\%) $\uparrow$ & 88.54 & 88.84 & 67.92 & 89.32 & 89.17 & 92.20 & \underline{93.67} & 89.24 & \cellcolor{gray!12}\textbf{98.24} \\
			PRO-AUC (\%) $\uparrow$   & 73.46 & 46.16 & 20.88 & 58.21 & 64.92 & 72.15 & \underline{76.27} & 72.28 & \cellcolor{gray!12}\textbf{89.88} \\
			\bottomrule
		\end{tabular}%
	}
	
	\vspace{2pt}
	\caption{\textbf{Comparison on Image-/Pixel-/PRO-AUC.} Best is in \textbf{bold}, second-best is \underline{underlined}. The publication venue (conference/journal) and year are shown under each method.}
	\label{tab:auc_comparison_with_venue}
\end{table*}

\begin{figure*}[t]
	\centering
	\includegraphics[width=1\linewidth]{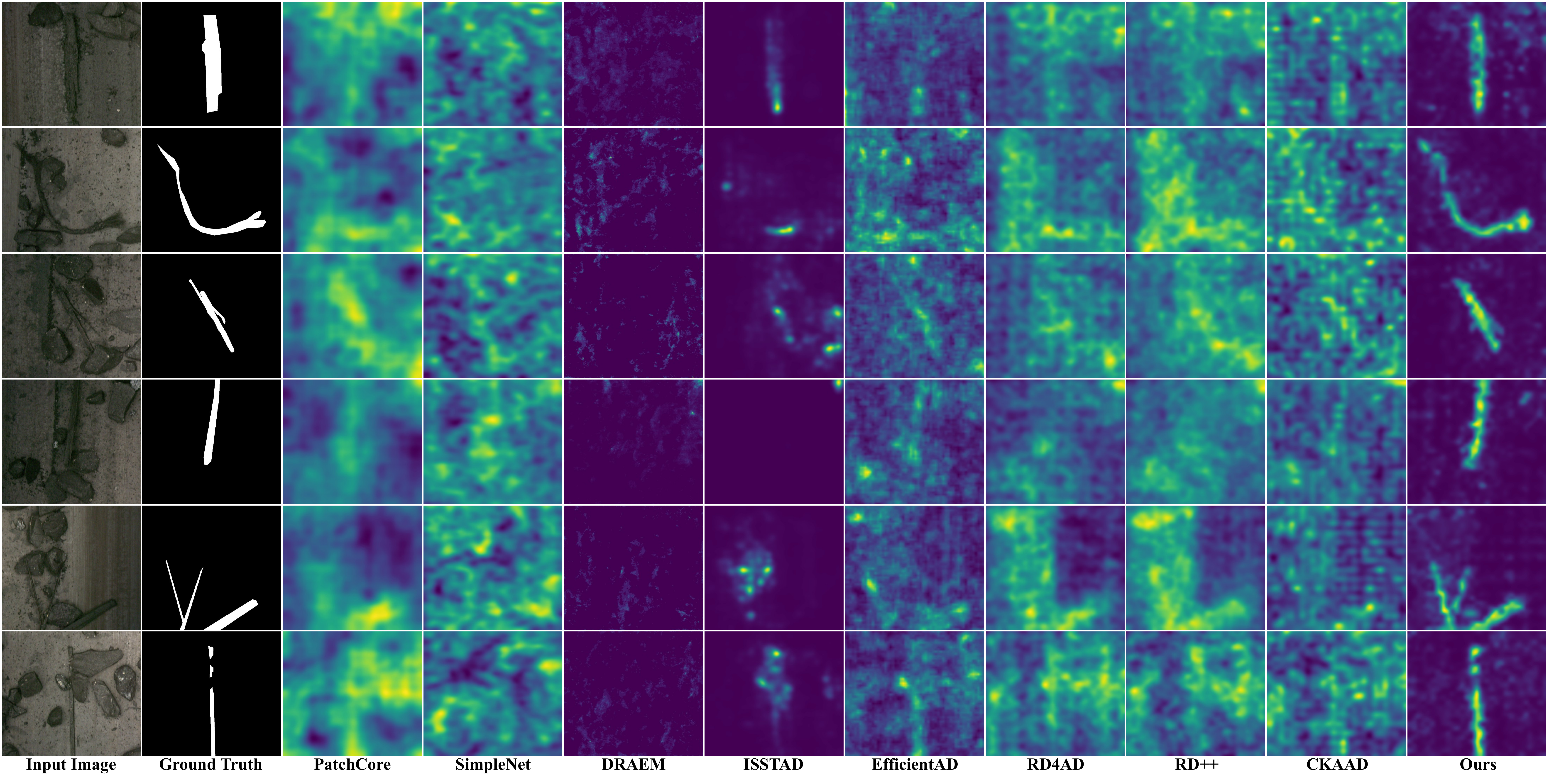}
	\caption{\textbf{Qualitative comparison on challenging (low-contrast) cases.} Pixel-level anomaly maps on CoalAD where foreign objects are heavily stained, compressed, or occluded and thus exhibit low contrast against the background. Most baselines fail to produce focused and coherent responses, while our method localizes the foreign-object regions with clearer and more complete contours.}
	\label{fig:Result_Hard}
\end{figure*}

\begin{figure*}[t]
	\centering
	\includegraphics[width=1\linewidth]{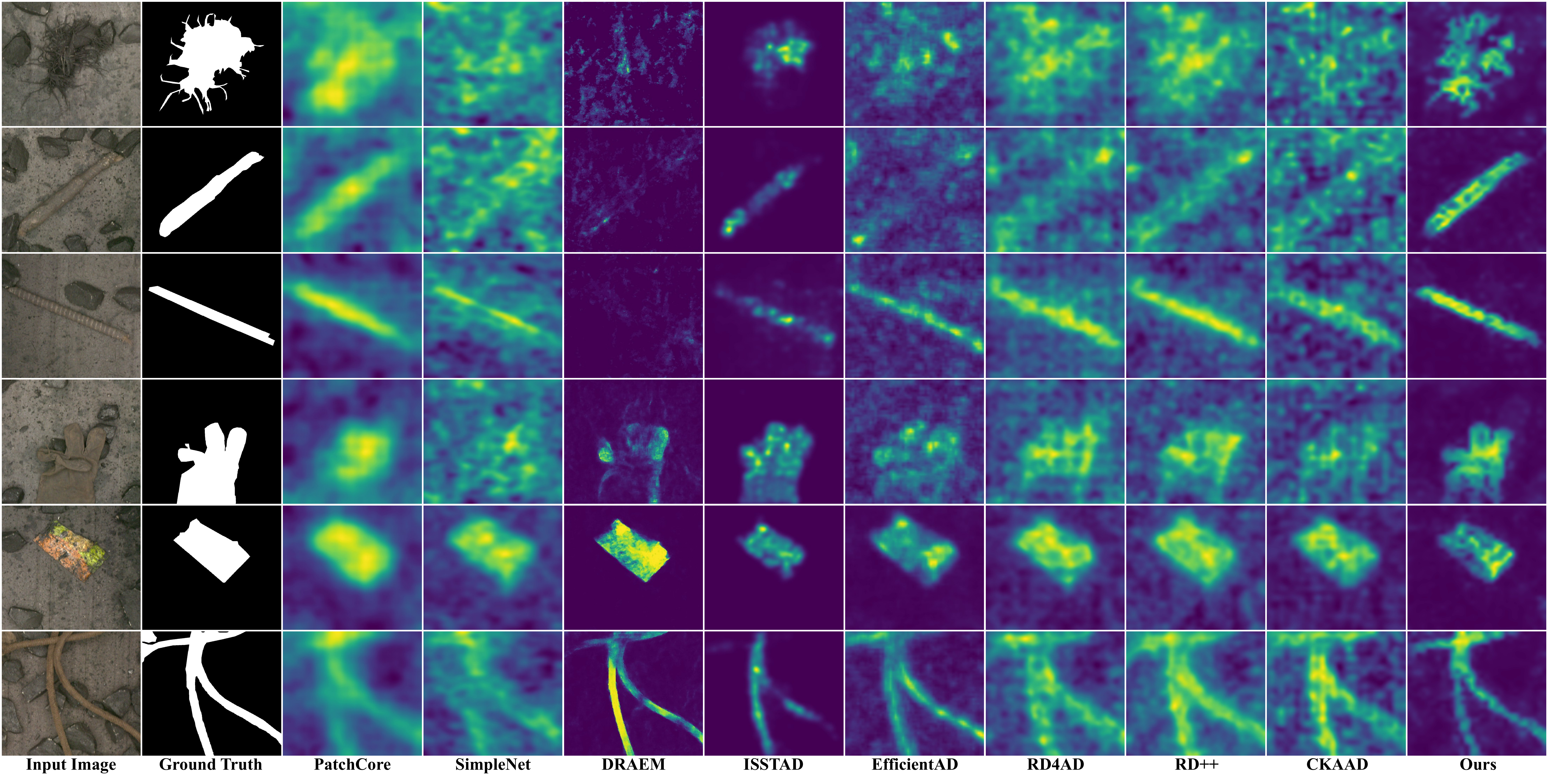}
	\caption{\textbf{Qualitative comparison on easier (high-contrast) cases.} Pixel-level anomaly maps on CoalAD where foreign objects show clear appearance differences from the background (e.g., color, shape, or boundary cues). Most methods can roughly indicate the object location, but may yield fragmented responses or spurious activations; our method provides the sharpest and most accurate localization.}
	\label{fig:Result_Simple}
\end{figure*}

\subsubsection{Baseline Methods}
We compare our method against a comprehensive set of widely adopted and representative baselines that cover major paradigms in industrial anomaly detection:
\begin{itemize}
	\item \textbf{Memory-based:} PatchCore~\cite{PatchCore}.
	\item \textbf{Reconstruction-based:} DRAEM~\cite{DRAEM}, ISSTAD~\cite{ISSTAD} (Transformer-based), CKAAD~\cite{CKAAD} (adversarially regularized).
	\item \textbf{Knowledge Distillation:} RD4AD~\cite{RD4AD}, RD++~\cite{RDplus}, EfficientAD~\cite{EfficientAD}.
	\item \textbf{Synthesis-driven:} SimpleNet~\cite{SimpleNet}.
\end{itemize}
All baselines are evaluated using their official implementations with recommended settings. For PatchCore, we use a widely adopted public re-implementation following the original paper and keep the core algorithm unchanged. All methods are evaluated under the same data split for a fair comparison.

\subsubsection{Quantitative Results}
The main quantitative comparison is presented in Table~\ref{tab:auc_comparison_with_venue}. A key observation is that most existing methods experience significant performance degradation on the highly \emph{non-structural} CoalAD scenario. Our method, in contrast, achieves state-of-the-art performance across all three metrics, demonstrating superior robustness.

\textbf{Analysis of Baseline Performance.}
The performance drop varies across paradigms, revealing their inherent limitations in unstructured environments:
\begin{itemize}
	\item \textbf{Reconstruction-based methods (DRAEM)} suffer the most severe degradation. Their effectiveness hinges on learning a stable distribution of normal appearances, an assumption frequently violated in coal-stream scenes with drastic variations and weak boundaries. While ISSTAD (Transformer backbone) and CKAAD (adversarial alignment) show some improvement, they remain fundamentally constrained by the reconstruction paradigm's sensitivity to unstructured noise.
	\item \textbf{Distillation-based methods (RD4AD, RD++, EfficientAD)} exhibit moderate degradation. Their reliance on feature-level consistency offers more robustness than pixel-level reconstruction. However, the student network's tendency to \emph{mimic} rather than \emph{understand} teacher features makes it difficult to amplify subtle semantic discrepancies when anomalies have low contrast with the background.
	\item \textbf{The synthesis-based method (SimpleNet)} shows relatively stable image-level detection, likely benefiting from its explicit training on discriminative, higher-level cues.
	\item \textbf{The memory-based method (PatchCore)} emerges as the strongest baseline. Its local patch matching mechanism can accommodate diverse local appearances, making it less reliant on global structural regularity. However, its lack of higher-level semantic constraints limits its ability to handle anomalies that are locally similar to normal texture or require semantic reasoning.
\end{itemize}

\subsubsection{Qualitative Results and Analysis}
Figs.~\ref{fig:Result_Hard} and \ref{fig:Result_Simple} provide qualitative insights, comparing pixel-level heatmaps on challenging low-contrast and relatively easier high-contrast cases, respectively.

\textbf{Low-Contrast Challenges (Fig.~\ref{fig:Result_Hard}):} When foreign objects are heavily stained, compressed, or occluded, most baselines produce weak, fragmented, or diffuse activations, failing to delineate object boundaries. Our method, leveraging consensus among complementary cues, delivers more coherent and complete localization.

\textbf{High-Contrast Cases (Fig.~\ref{fig:Result_Simple}):} Even when objects are more visually distinct, baseline responses can be noisy, over-smoothed, or imprecise. Our method consistently provides the sharpest and most accurate spatial localization, highlighting its precision.

\subsubsection{Summary and Discussion}
The experimental results collectively underscore the challenge posed by unstructured conveyor-belt coal scenes, where unstable local textures and complex backgrounds break the core assumptions of many existing paradigms. Methods over-reliant on stable reconstruction, pure local similarity, or feature imitation struggle. Our multi-branch framework addresses these limitations by synergistically combining object-level semantic modeling, global semantic attribution, and fine-grained texture matching, thereby achieving robust and precise anomaly perception in this demanding environment.

\subsection{Ablation Studies}
\label{sec:ablation}

\subsubsection{Contribution of Individual Branches and Fusion Strategy}
To validate the design of our multi-branch framework, we systematically quantify the contribution of each branch to image-level detection and pixel-level localization, and evaluate the effectiveness of our fusion strategy.

\textbf{Image-Level Detection.}
We compare three sources of image-level anomaly evidence: the global semantic deviation score \(s_{\text{obj}}\) from the object-level branch, the local texture anomaly score \(s_{\text{pc}}\) from the texture branch, and the map-pooled score \(s_{M}\) derived from the fused pixel-level heatmap \(M_{\text{fused}}\) (by taking its maximum value). We then evaluate their pairwise linear combinations and the final fused score \(S_{\text{fused}}\) defined in Eq.~\eqref{eq:image_fusion_final}. As shown in Table~\ref{tab:ablation_image_fusion}, any pairwise fusion outperforms the corresponding single scores, and the full fusion achieves the best image-level AUC. This demonstrates the complementary nature of these cues: \(s_{\text{obj}}\) captures deviations in global object composition, \(s_{\text{pc}}\) is sensitive to local texture irregularities, and \(s_{M}\) provides a spatially aware signal that reflects the strongest localized anomaly after pixel-level consensus.

\textbf{Pixel-Level Localization.}
We evaluate the three branch-specific anomaly maps—\(M_{\text{obj}}\) (object-level), \(M_{\text{attr}}\) (semantic attribution), and \(M_{\text{pc}}\) (texture)—along with their pairwise Hadamard (\(\odot\)) products and the final fused map \(M_{\text{fused}}\) (Eq.~\eqref{eq:pixel_fusion}). Table~\ref{tab:ablation_pixel_fusion} shows that all pairwise fusions yield clear improvements over any single map, confirming the complementary strengths of each branch. The object-level map provides stable semantic constraints, the attribution map highlights regions responsible for global semantic deviation, and the texture map captures fine-grained local details. Their fusion effectively suppresses spurious activations from background noise while amplifying true foreign-object regions. The full three-branch fusion achieves the best pixel-level AUC and PRO-AUC, indicating that all three cues are non-redundant and synergistically enhance localization robustness and boundary accuracy in this non-structured scenario.

\begin{figure*}[t]
	\centering
	\includegraphics[width=1\linewidth]{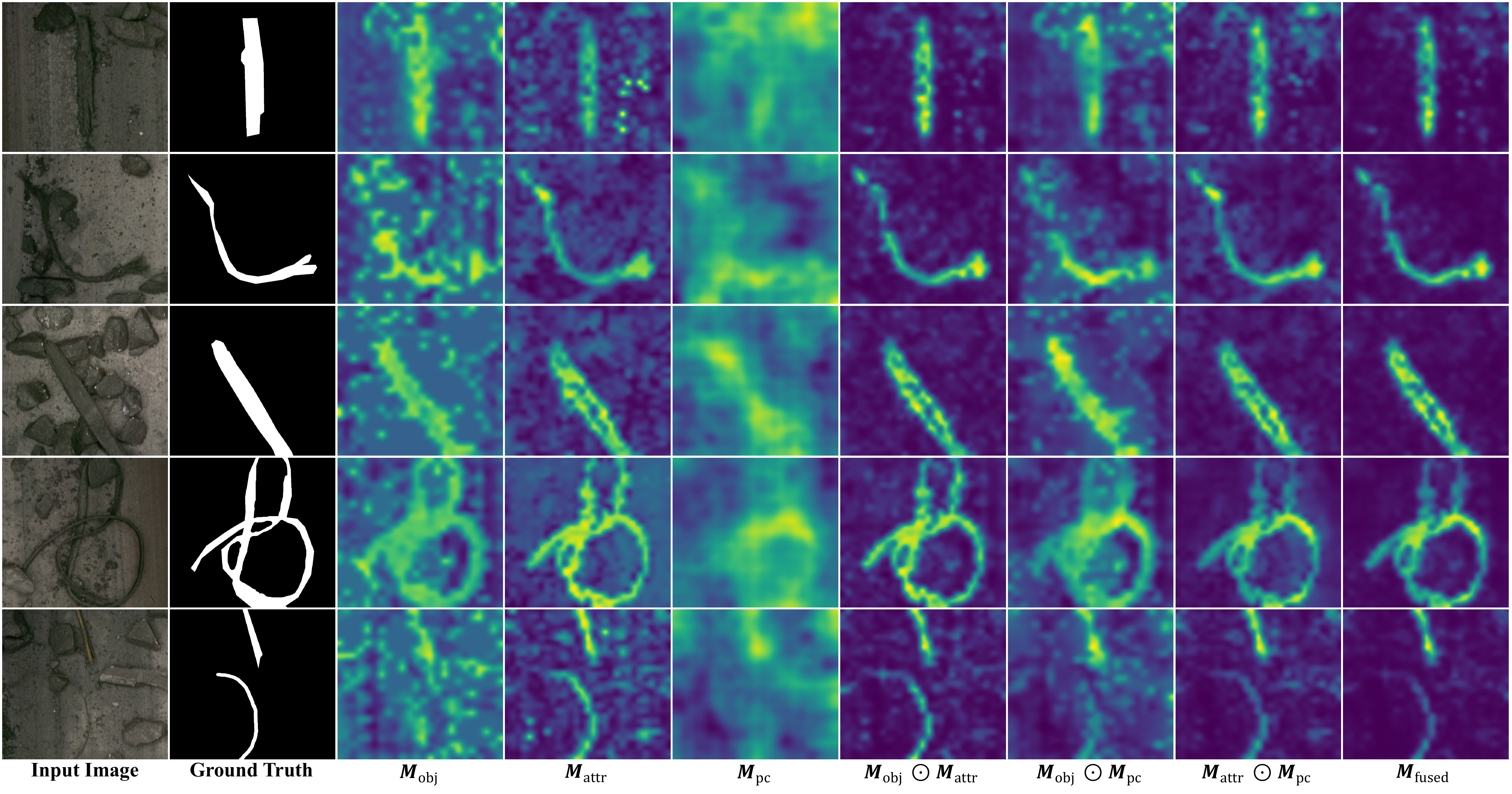}
	\caption{\textbf{Visualization of pixel-level fusion.} Pixel-level anomaly score maps from individual branches, pairwise fusion, and the final three-branch fusion. Compared with single-branch results, fusion produces more reliable localization, and incorporating all three branches further improves the overall quality.}
	\label{fig:Result_Fuion}
\end{figure*}

\textbf{Qualitative Visualization.}
Fig.~\ref{fig:Result_Fuion} provides a visual comparison of the heatmaps. Single-branch maps often exhibit noise, fragmentation, or incomplete coverage. Pairwise fusion yields more focused and coherent responses, and the full three-branch fusion produces the sharpest, most complete contours that best align with the ground-truth masks, corroborating the quantitative findings.

\begin{table*}[t]
	\centering
	\small
	\setlength{\tabcolsep}{5.6pt}
	\renewcommand{\arraystretch}{1.15}
	
	\resizebox{\textwidth}{!}{%
		\begin{tabular}{lccccccc}
			\toprule
			& \(s_{\text{obj}}\) & \(s_{\text{pc}}\) & \(s_{M}\)
			& \(s_{\text{obj}}{+}s_{\text{pc}}\) & \(s_{\text{obj}}{+}s_{M}\) & \(s_{\text{pc}}{+}s_{M}\)
			& \(\mathbf{S}_{\text{fused}}\) \\
			\midrule
			Image-AUC (\%) \(\uparrow\)
			& 97.32 & 94.27 & 94.48
			& 98.52 & 98.19 & 96.91
			& \textbf{98.83} \\
			\bottomrule
		\end{tabular}%
	}
	
	\vspace{2pt}
	\caption{\textbf{Ablation on image-level detection.} Performance of individual image-level scores, pairwise fusion, and the final fused score.}
	\label{tab:ablation_image_fusion}
\end{table*}

\begin{table*}[t]
	\centering
	\small
	\setlength{\tabcolsep}{4.8pt}
	\renewcommand{\arraystretch}{1.15}
	
	\resizebox{\textwidth}{!}{%
		\begin{tabular}{lccccccc}
			\toprule
			& \(M_{\text{obj}}\) & \(M_{\text{attr}}\) & \(M_{\text{pc}}\)
			& \(M_{\text{obj}}\odot M_{\text{attr}}\) & \(M_{\text{obj}}\odot M_{\text{pc}}\) & \(M_{\text{attr}}\odot M_{\text{pc}}\)
			& \(\mathbf{M}_{\text{fused}}\) \\
			\midrule
			Pixel-AUC (\%) \(\uparrow\)
			& 95.75 & 94.08 & 88.42
			& 97.43 & 96.96 & 96.90
			& \textbf{98.24} \\
			PRO-AUC (\%) \(\uparrow\)
			& 81.15 & 75.90 & 74.65
			& 86.96 & 87.36 & 85.05
			& \textbf{89.88} \\
			\bottomrule
		\end{tabular}%
	}
	
	\vspace{2pt}
	\caption{\textbf{Ablation on pixel-level localization.} Performance of individual anomaly maps, pairwise fusion by Hadamard product (\(\odot\)), and the final fused map.}
	\label{tab:ablation_pixel_fusion}
\end{table*}

\subsubsection{Sensitivity to Key Hyperparameters}
We analyze the sensitivity of our framework to two critical hyperparameters: the input resolution for the ViT-based branches and the foreground suppression threshold \(\tau\) in the object-level branch.

\textbf{Input Resolution for ViT-based Branches.}
The input resolution determines the granularity of patch tokens and the computational cost. We perform a resolution sweep for both ViT-based branches.

\begin{figure}
	\centering
	\includegraphics[width=1\linewidth]{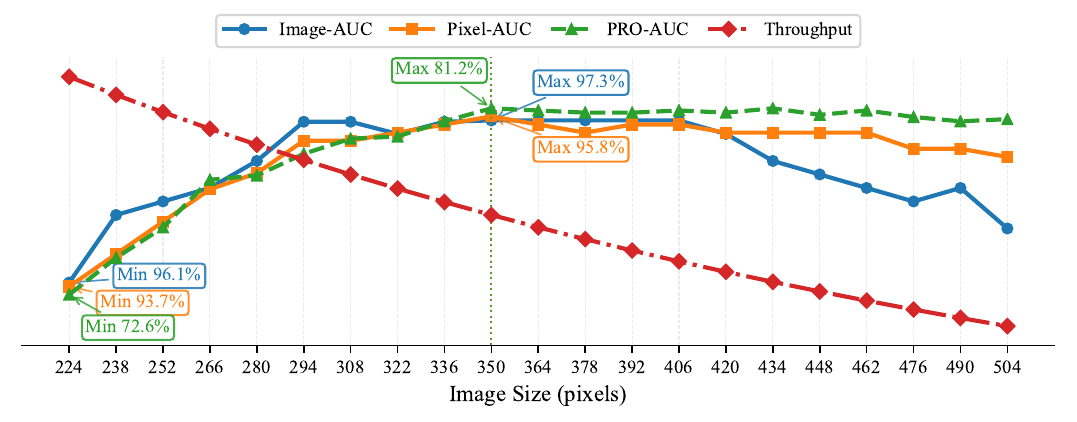}
	\caption{\textbf{Resolution ablation for the object-level branch (MACLU).} Metrics and throughput versus input resolution. AUCs peak around \(350\times350\), which is chosen as the default.}
	\label{fig:maclu_size}
\end{figure}

\begin{figure}
	\centering
	\includegraphics[width=1\linewidth]{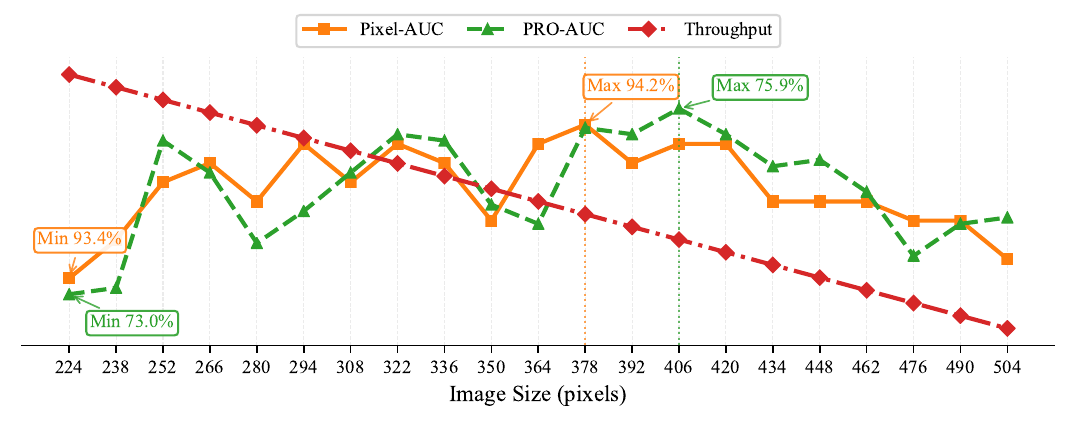}
	\caption{\textbf{Resolution ablation for the semantic attribution branch (MA2Patch).} Pixel-AUC, PRO-AUC, and throughput versus input resolution. PRO-AUC is maximized at \(406\times406\).}
	\label{fig:m2p_size}
\end{figure}

\begin{itemize}
	\item \textbf{Object-level Branch (MACLU):} Fig.~\ref{fig:maclu_size} shows the trade-off between accuracy and throughput as resolution varies from \(224\times224\) to \(504\times504\). The three AUC metrics peak or plateau around \(350\times350\), while throughput declines with higher resolution. We therefore set the default resolution to \(350\times350\).
	\item \textbf{Semantic Attribution Branch (MA2Patch):} Fig.~\ref{fig:m2p_size} indicates that Pixel-AUC remains stable in a wide range (\(\sim 304\)–\(420\)), while PRO-AUC is maximized at \(406\times406\). We select \(406\times406\) as the default, balancing localization quality and efficiency.
\end{itemize}

\begin{figure}
	\centering
	\includegraphics[width=1\linewidth]{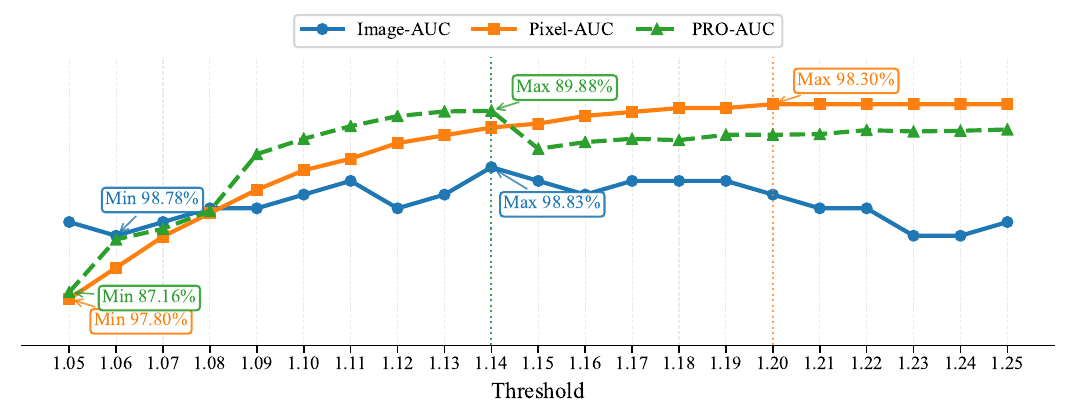}
	\caption{\textbf{Sensitivity of the foreground suppression threshold \(\tau\) in MACLU.} Final fused performance versus \(\tau\). Image-AUC and PRO-AUC peak around \(\tau=1.14\).}
	\label{fig:replace_thr}
\end{figure}

\textbf{Foreground Suppression Threshold \(\tau\).}
In the object-level branch, the threshold \(\tau\) controls the suppression of strong foreground (coal/gangue) patches via mean replacement. Fig.~\ref{fig:replace_thr} shows the impact of \(\tau\) (swept from \(1.05\) to \(1.25\)) on the final fused performance. As \(\tau\) increases, all metrics initially improve, indicating that moderate suppression stabilizes anomaly responses by reducing interference from extreme foreground patches. Image-AUC and PRO-AUC peak around \(\tau = 1.14\), after which they decline, suggesting that an overly large threshold may fail to cover extreme foreground variations, leading to mis-localization. Pixel-AUC continues to rise slightly but marginally. Considering the overall detection-localization trade-off, we set \(\tau = 1.14\).

\subsection{Failure Cases}

\begin{figure}[t]
	\centering
	\includegraphics[width=1\linewidth]{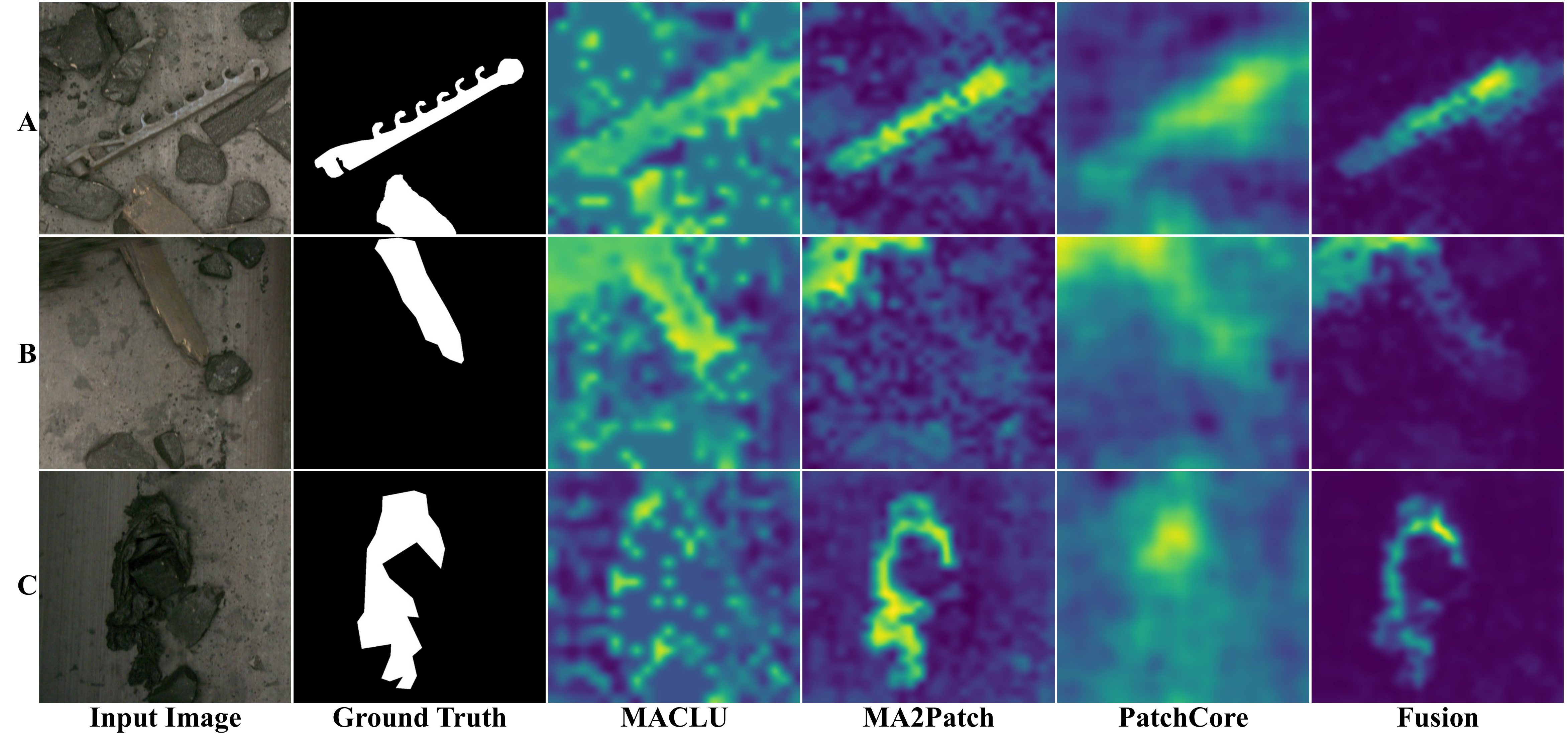}
	\caption{\textbf{Representative failure cases of our method.} Three typical scenes where our multi-branch fusion fails to produce accurate localization, illustrating the remaining challenges in highly unstructured coal-stream scenes.}
	\label{fig:failure_cases}
\end{figure}
While our method demonstrates strong overall performance, analysis of its failure cases provides crucial insights into remaining challenges and limitations. Fig.~\ref{fig:failure_cases} presents three representative scenes where the fused output is incorrect or incomplete. By examining the responses of individual branches, we can diagnose the underlying causes.

\textbf{Case A.}
The final fused map highlights only the metallic foreign object while missing the other wooden one in the same scene. By inspecting the localization results of the three branches, we find that the main cause is that the semantic-attribution branch responds much more strongly to the metal object---whose shape and material deviate more drastically from normal samples than the wooden one. As a result, this branch concentrates most anomaly evidence on the metal object and downplays the wooden object. Meanwhile, the other two branches produce only weak responses on the wooden object, which is insufficient to correct the semantic branch's tendency to treat it as background during fusion, leading to a miss.

\textbf{Case B.}
This scene contains camera-induced artifacts. Similar to Case A, the semantic-attribution branch reacts strongly to the disturbed region (i.e., high semantic abnormality) and thus overlooks the true foreign object, which eventually causes an incorrect fused prediction. This case suggests that salient non-target disturbances such as noise, streaks, or local exposure artifacts can mislead semantic-level anomaly evidence and dominate the fusion outcome.

\textbf{Case C.}
The foreign object is highly deceptive, with appearance patterns that are strongly confusable with the background texture/structure. Consequently, both the object-level branch and the texture branch show little to no response. Although the semantic-attribution branch detects the anomalous region to some extent, the weak evidence from the other branches suppresses this response in the fusion stage, resulting in an under-activated final localization. This case indicates that when anomalies are highly similar to the background in terms of local texture and object composition, anomaly evidence from a single branch may not be preserved or amplified through fusion.

\section{Conclusion}

We study unsupervised foreign-object anomaly detection and pixel-level localization in highly unstructured coal-stream conveyor scenes, where unstable normal appearances and low-contrast, heavily disturbed foreign objects weaken the assumptions behind many methods designed for structured industrial settings. To support evaluation in this scenario, we introduce \textbf{CoalAD}, a benchmark dataset with a standardized protocol for image-level detection and pixel-level localization.

We further propose a fusion framework that integrates complementary cues from three branches: object-level semantic composition modeling (MACLU), semantic-attribution-based global deviation analysis (MA2Patch), and fine-grained texture matching (PatchCore). Experiments on CoalAD show consistent improvements over widely adopted representative baselines, and ablations verify the contribution of each component. Failure cases indicate that branch dominance and weak cross-branch agreement can both degrade fusion, motivating more adaptive, confidence-aware fusion.

\bibliographystyle{unsrt}  
\bibliography{refs}

\end{document}